\documentclass{class/ieeeaccess}
\usepackage{cite}
\usepackage{amsmath,amssymb,amsfonts}
\usepackage{algorithmic}
\usepackage{textcomp}
\usepackage{tabularx}
\usepackage{booktabs}	
\usepackage{soul}
\usepackage{pifont}
\usepackage{url}
            
\ifCLASSINFOpdf
   \usepackage[pdftex]{graphicx}
   \graphicspath{{figures/}}
   \DeclareGraphicsExtensions{.pdf,.jpeg,.png}
\else
   \usepackage[dvips]{graphicx}
   \graphicspath{{figures/}}
   \DeclareGraphicsExtensions{.eps}
\fi

\newcolumntype{P}[1]{>{\centering\arraybackslash}p{#1}}

\begin{document}
\history{Received 10 August 2023, accepted 2 September 2023, date of publication 6 September 2023, \\
date of current version 12 September, 2023.}
\doi{10.1109/ACCESS.2023.3312382}

\title{Radars for Autonomous Driving: A Review of Deep Learning Methods and Challenges}
\author{
\uppercase{Arvind Srivastav}\authorrefmark{1} 
and
\uppercase{Soumyajit Mandal}\authorrefmark{2}
\IEEEmembership{Senior Member, IEEE}}
\address[1]{Zoox, Inc., Foster City, CA 94404}
\address[2]{Brookhaven National Laboratory, Upton, NY 11973}


\markboth
{A. Srivastav, S. Mandal: Radars for Autonomous Driving: A Review}
{A. Srivastav, S. Mandal: Radars for Autonomous Driving: A Review}

\corresp{Corresponding author: Arvind Srivastav (e-mail: arvindsr33@gmail.com)}

\begin{abstract}
Radar is a key component of the suite of perception sensors used for safe and reliable navigation of autonomous vehicles. Its unique capabilities include high-resolution velocity imaging, detection of agents in occlusion and over long ranges, and robust performance in adverse weather conditions. However, the usage of radar data presents some challenges: it is characterized by low resolution, sparsity, clutter, high uncertainty, and lack of good datasets. These challenges have limited radar deep learning research. As a result, current radar models are often influenced by lidar and vision models, which are focused on optical features that are relatively weak in radar data, thus resulting in under-utilization of radar's capabilities and diminishing its contribution to autonomous perception. This review seeks to encourage further deep learning research on autonomous radar data by 1) identifying key research themes, and 2) offering a comprehensive overview of current opportunities and challenges in the field. Topics covered include early and late fusion, occupancy flow estimation, uncertainty modeling, and multipath detection. The paper also discusses radar fundamentals and data representation, presents a curated list of recent radar datasets, and reviews state-of-the-art lidar and vision models relevant for radar research. 

\end{abstract}

\begin{keywords}
Radar, perception, autonomous driving, self-driving cars, electric vehicles, 4D data, deep learning, early fusion, occupancy estimation, transformers, point cloud, uncertainty modeling, multipath 
\end{keywords}

\titlepgskip=-15pt

\maketitle
\section{Introduction}
\label{sec:introduction}
\IEEEPARstart{M}{odern} electric and hybrid vehicles are increasingly targeting advanced levels of autonomy, from hands-free driving to full self-driving. These levels of autonomy require an ego vehicle to safely navigate through complex driving conditions that include dense traffic, agents incoming from occlusion and long ranges, and adverse weather. Agents within the scene vary extensively, from common entities such as cars, bikes, and pedestrians, to rarer types like flatbed trucks, construction vehicles, airborne chairs, and rolling tires. The perception system is responsible for detection and tracking of these agents using a suite of sensors -- cameras, radars, and often, lidars. These sensors image the environment and run deep learning models on the collected data to enable continuous operation.

\Figure[t!](topskip=0pt, botskip=0pt, midskip=0pt)[width=\textwidth]{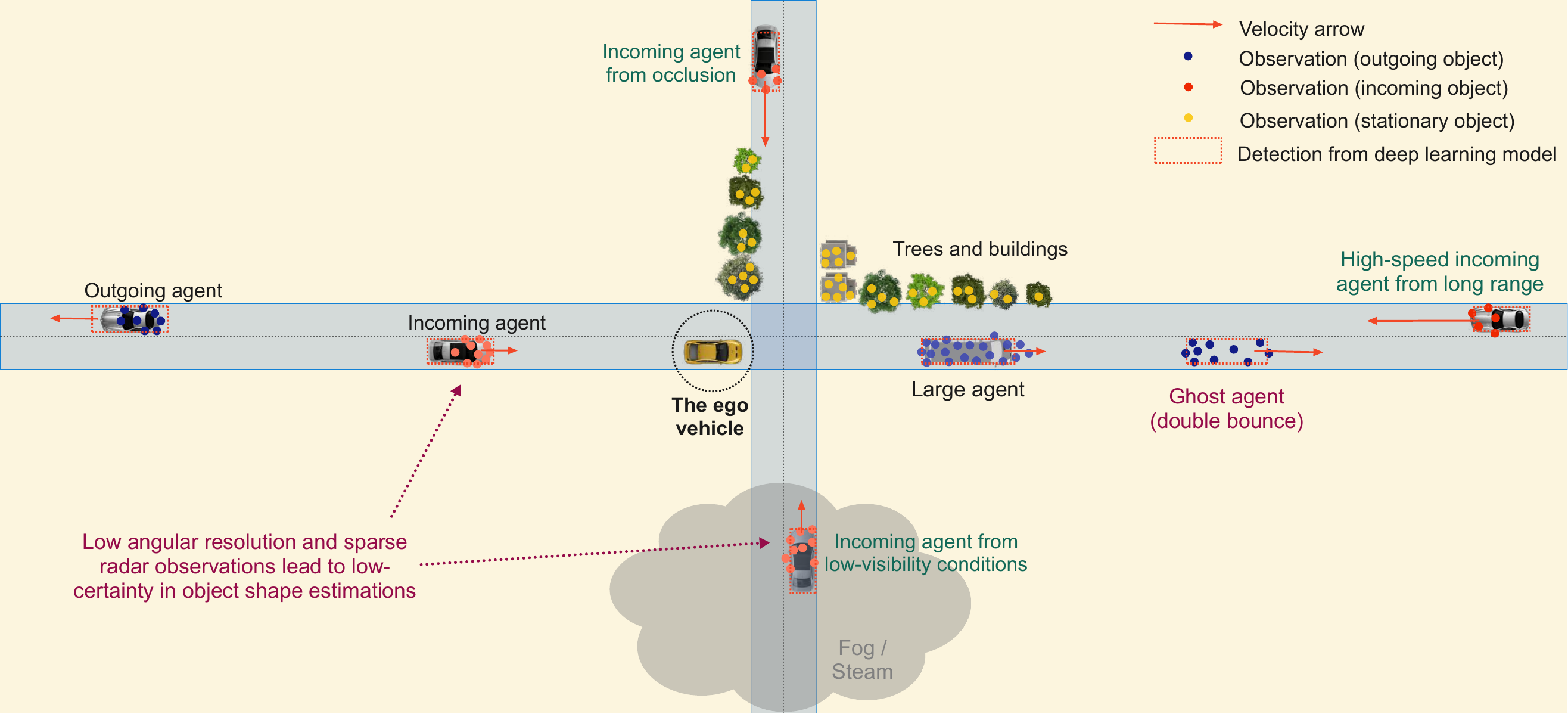}{An overview of the capabilities and challenges associated with using radars for autonomous driving. Radars excel in getting observations from occluded agents, covering long ranges, and performing in adverse weather conditions. However, their data is often sparse, low resolution, contains clutter, and has high uncertainty. These issues can lead to low confidence in shape estimation and false positive detection by deep learning models.\label{fig:radar-applications}}%

Camera and lidar are visible-spectrum sensors which offer rich scene information but face limitations. Cameras provide rich semantic details of the scene, but they are affected by lighting and weather conditions and lack depth information vital for object tracking. Lidars provide dense and accurate depth information, but suffer from scattering due to weather~\cite{Bijelic2018ABF}. 
 
By contrast, radars provide high-resolution range and velocity imaging and excel at detecting agents early in occlusion and over long ranges due to their broad beam and high receiver sensitivity~\cite{MOTRadar}. Radars can observe agents up to a few seconds before cameras or lidars, a crucial advantage in situations requiring swift response from the ego vehicle. Moreover, radars undergo little scattering in rain, fog and dust clouds due to their relatively large wavelength (millimeters)~\cite{Zhang2021PerceptionAS, Dickmann2016AutomotiveRT}, making them tolerant to weather conditions. Fig.~\ref{fig:radar-applications} illustrates some important capabilities of radars used in autonomy. 

However, radar data comes with challenges. Angle resolution is relatively low, resulting in sparse data and significant uncertainty in the position of agents, especially at long ranges~\cite{Patel2021InvestigationOU}. The large wavelength makes nearby surfaces behave like mirrors, leading to multipath propagation that adds clutter -- false positive observations and noise -- in radar data~\cite{Prophet2019InstantaneousGD}.   
 
Detection and tracking of agents in these complex scenes requires deep learning models. As with vision and lidar research, training these models requires quality radar datasets which were scarce until recently. The first acceptable radar dataset, NuScenes~\cite{Caesar2019nuScenesAM}, was released only in 2019, and higher quality datasets are just now becoming available~\cite{Paek2022KRadar4R, Palffy2022MulticlassRU, Schumann2021RadarScenesAR, burnett2023boreas}. This data shortage, combined with the challenges of radar data, have led to a dearth of radar-focused models~\cite{Zhou2022TowardsDR, Venon2022MillimeterWF, Liu2021DeepIS, Yang2020RadarNetER, Palffy2020CNNBR}. 

As a result, current radar models are often based on lidar and camera detection models~\cite{Lang2018PointPillarsFE, radarPointSemseg, Wang2021MultiModal3O, Abdu2021ApplicationOD} Such models are easily implemented after formatting radar data to match the required input format, such as a pseudo-image~\cite{Lang2018PointPillarsFE}. Additionally, perception has conventionally followed a late fusion-based multi-object tracking paradigm~\cite{Zhang2021ByteTrackMT} that needs full object detection from each sensor model. Employing lidar and camera models simplifies the task of radar since these models are designed for object detection. 
 
Lidar and camera models are primarily optimized to learn from optical features and produce outputs that include accurate class, shape, and orientation of objects. However, radars have relatively weak optical features, as evident from their low resolution, sparse, clutter-filled, and highly uncertain data. Instead, radars offer unique features like high resolution range and velocity imaging and early detection. Employing camera and lidar detection models therefore under-utilizes the capabilities of radar, leading to less reliable detection and reduced contribution of radars to perception tasks. 
 
Recently, researchers have begun exploring new types of models to enhance the contribution of radar to perception, including early fusion models, occupancy estimation models, and uncertainty-based models. The ``late fusion'' approach traditionally used for perception tasks requires full object detection from each sensor's data and lacks access to sensor data features during fusion. This approach makes the final fused outputs less robust. By contrast, early fusion models learn jointly from multi-sensor data features to produce robust unified detection results. Early fusion has gained more traction with the advent of transformers~\cite{vaswani2017attention}, which use cross-attention to learn effectively from dissimilar modalities. Radar-based early fusion models primarily use an optical sensor data for class, shape, and orientation inferences, and radar data to enhance depth and velocity estimations~\cite{Yao2023RadarCameraFF, Tang2021OnRoadOD, Farag2021RealtimeLA, Liu2021SurroundingOD}. Certain radar-based early fusion models also achieve tolerance against weather or lighting conditions by leveraging radar features in innovative architectures and training methods~\cite{Li2022ModalityAgnosticLF, Hwang2022CramNetCF}. 
 
Both late fusion and early fusion paradigms in perception require object detection for tracking. Detection models, however, can be unpredictable on rare types of agents like an airborne chair or rolling tyre due to their under-representation in the training dataset. Thus, these rare cases make perception susceptible to failure. 
 
To overcome this issue, occupancy estimation models dispense with the need for detection during tracking. These models divide the scene into high-resolution 2D or 3D grids and learn occupancy and velocity features for each cell~\cite{tesla_occupancy}. They track these features over frames to estimate the drivable area for ego vehicle navigation. Using this approach, occupancy models achieve robustness to all shapes and types of agents. Radar-based occupancy~\cite{Hoermann2017DynamicOG, Nuss2016ARF} and scene flow~\cite{Menze2017ObjectSF, Mittal2019JustGW, Pontes2020SceneFF} estimation models contribute more significantly to this paradigm by providing features rich in velocity, range, and early detection that are tolerant to weather conditions. 

As previously mentioned, radar data is far from perfect. The challenges for a radar model increase when considering that object labels for radar data are typically copied from lidar and camera labels~\cite{Sheeny2020RADIATEAR}, which may not be very accurate for the radar data. Inaccuracies can arise in labels due to issues with sensor time-synchronization, the low angular resolution of radar, and differences in sensor imaging physics. Additionally, labels for all objects in the dataset carry the same confidence, irrespective of their distance, size, or visibility. Training models on such a dataset frequently produces incorrect and overconfident detections as the data doesn't capture the inherent uncertainty in labels~\cite{Gal2016UncertaintyID}. Some studies leverage the uncertainty in the data and labels~\cite{Patel2021ImprovingUO} to train probabilistic models~\cite{Feng2020ARA} that improve detection performance for low-certainty agents. 

Radars used in autonomy generally belong to one of two generations. Most are older-generation radars, referred to as \textit{3D radars}, that image the scene in three dimensions -- range, velocity, and azimuth angle. These radars provide data that is sparse and low-resolution and lacks height information. On the other hand, a few newer-generation radars, termed \textit{4D radars}, markedly enhance density and resolution compared to 3D radars and provide an elevation angle dimension. These devices are more costly, but their data holds the potential to significantly boost the performance of learned radar models due to stronger features. However, research using superior quality 4D radar datasets, such as the K-radar~\cite{Paek2022KRadar4R}, is still limited, largely due to the prevalence of the 3D radar NuScenes dataset as the benchmark for comparison. We anticipate that the adoption of 4D radars will accelerate as researchers begin to publish their benchmark results on 4D radar datasets. 

In this review, we examine state-of-the-art radar models for the areas identified above and discuss challenges and opportunities within these areas. Additionally, we discuss the fundamentals of radar, explore radar data representation, and provide a curated list of both 3D and 4D autonomous radar datasets. Further, as 4D radars make strides towards enhancing the quality of optical-like features, we also review relevant state-of-the-art lidar and vision models that may guide research on generation and training of radar models.  

The rest of the paper is organized as follows: Section~\ref{sec:radar_overview} offers an intuitive understanding of radar fundamentals and explores various radar data formats and their representations for deep learning. Section~\ref{sec:datasets} introduces a curated list of recent radar datasets useful for radar deep learning research. Section~\ref{sec:pcp-overview} provides an overview of perception and the integration of different areas of radar deep learning within perceptual models. Section~\ref{sec:detection-based-tracking} reviews state-of-the-art radar models in both late fusion and early fusion paradigms. Section~\ref{sec:occupancy} reviews radar-only and multi-sensor fusion models proposed for occupancy and scene flow estimations. Section~\ref{sec:uncertainty} discusses uncertainty issues in radars used for autonomous navigation and discusses methods to reduce uncertainty in both the models and labels to improve detection accuracy. Section~\ref{sec:challenges} reviews research on reducing multipath clutter in radar data and discusses the prospects of generating synthetic radar data. The review concludes with a summary and our final thoughts.      
    
\section{Radar Fundamentals}
\label{sec:radar_overview}

\subsection{A brief history of radar}
Radars were invented during World War II primarily for detection and tracking of enemy aircraft from distances of hundreds of kilometers. By the late 20th century, radars had also found numerous applications in civilian sectors such as air traffic control and weather prediction. 

As technology advanced, radars evolved into lower-power, higher-resolution devices, opening up new applications, notably in the automotive sector around the turn of the 21st century. These automotive radars mainly support the advanced driver-assistance system (ADAS) in vehicles, enabling functionalities like adaptive cruise control and blind spot detection through classical radar signal processing algorithms.

The advent of the 2005 Darpa Grand Challenge~\cite{DARPA}  and AlexNet~\cite{Krizhevsky2012ImageNetCW} catalyzed the development of autonomous vehicles and associated deep learning research. Given the complexity of driving conditions, autonomous driving requires accurate instantaneous velocity estimations, navigation in adverse weather, and long-range detection. These needs have driven the development of autonomous radar-based sensors, including 3D radars, and more recently, 4D radars. These devices offer a substantial improvement in resolution compared to their automotive counterparts and facilitate the application of deep learning models to their data for detection and tracking in complex environments. 

\Figure[]()[page=9, width=\textwidth]{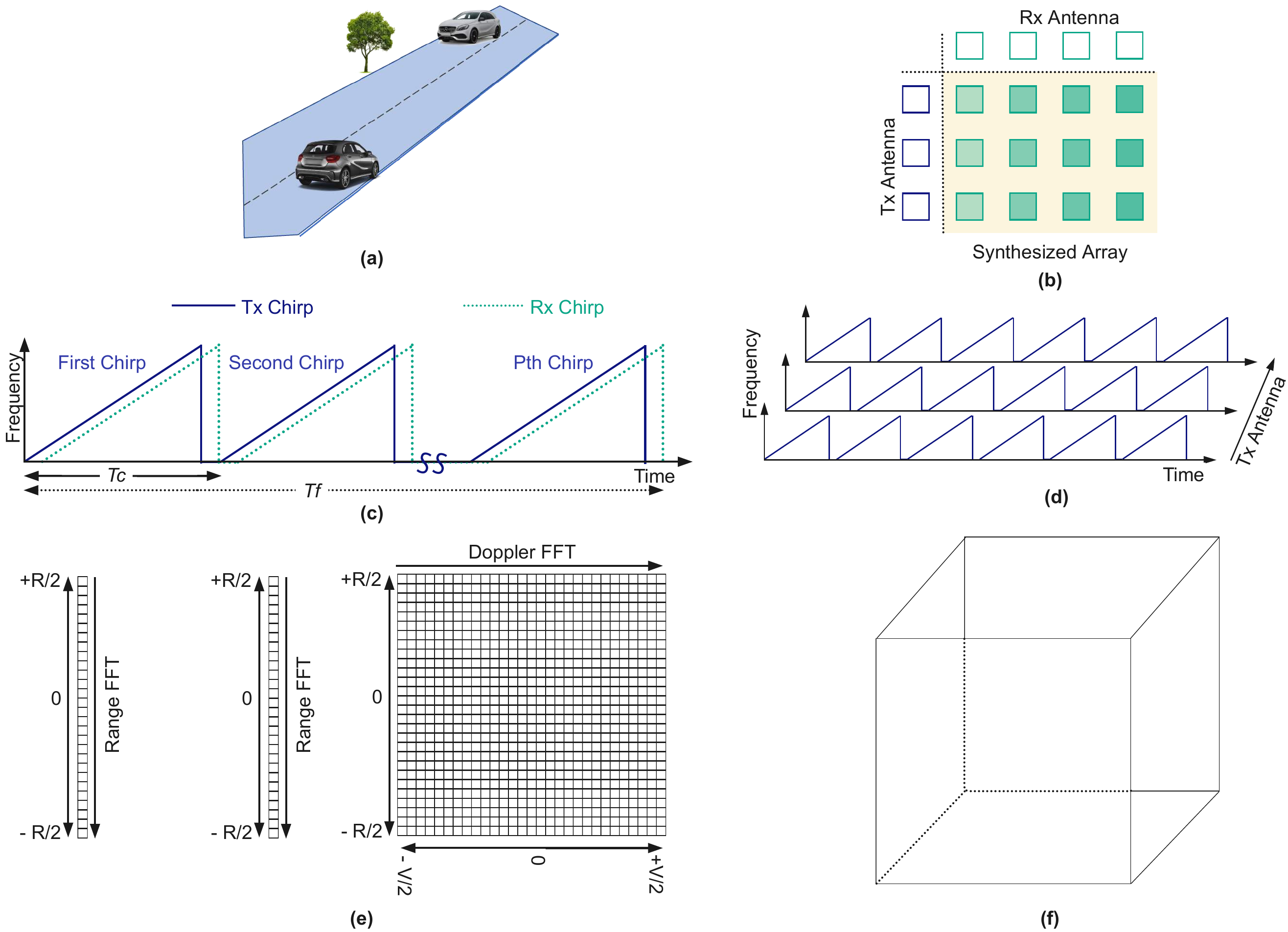}
{An illustration of the working of an FMCW radar. (a) An autonomous driving scene featuring two objects along with their relative ranges, velocities, and azimuth angles. (b) An FMCW radar antenna array composed of 3 transmitting antennas and 4 receiving antennas, forming a $3 \times 4$ array. (c) An FMCW frame with $P$ transmitted chirps and their corresponding received chirps; and (d) chirps from all three transmitters. (e) The range fast Fourier transform (FFT) of each chirp leading to a 2D range-Doppler spectrogram after performing a Doppler FFT. (f) The final radarcube obtained after performing an angle FFT.\label{fig:radar-basics}}


\subsection{Radar physics}
A radar is a time-of-flight sensor which provides an X-ray-like view by measuring the range, radial velocity, and angle of agents within its scene. It transmits electromagnetic (EM) waves within a solid angle at regular intervals, 
and for each transmitted wave, measures a \textit{range} of agents by computing the delay in receiving the wave reflected from those agents. By taking rapid successive range measurements from waves at regular intervals, it can also compute the rate of range change, or the \textit{radial velocity} of agents.

Radars use arrays of transmitter and receiver antennas to measure the angles of agents. For a transmitted EM wave, each receiver receives signals at varying delays from the same agent due to different path lengths created by the agent's angle relative to the transmitter orientation. These differences in delay assist in measuring the angles of agents in the scene. The \textit{azimuth angle} is determined by a horizontal array of antennas, while the \textit{elevation angle} is determined by a vertical array of antennas. The total size of the array in both dimensions (i.e., the spatial aperture) governs the diffraction-limited angular resolution. 

Nonetheless, due to physical constraints on the number of antennas that can be integrated into a standard-sized radar sensor, the angle measurement in radars used for autonomous navigation tends to be coarse, with a resolution generally exceeding $1^{\circ}$. Fig. \ref{fig:radar-basics}(a) illustrates the range ($R$), azimuth ($Az$), and radial velocity ($V_{rel}$) components of a radar observation from an incoming car.

\subsection{Overview of radars used in autonomous navigation} 
The radars used in autonomous navigation typically operate in the 77-81~GHz frequency band, as allocated by the Federal Communications Commission (FCC). The radar waves in this band have wavelength of $\sim$4~mm, which is larger than fog and dust particles and light rain droplets. Thus, these waves undergo minimal scattering, which makes them immune to adverse weather conditions. These radars provide high-resolution range and velocity imaging and low-resolution angle imaging of the scene and can observe agents early in occlusion and over long ranges. These characteristics, combined with immunity to weather conditions, make radars useful complementary sensors for autonomous perception. 

Radars used in autonomous navigation are primarily of two types: frequency-modulated continuous wave (FMCW)~\cite{SPatole} and coded~\cite{uhnderUhnderLaunches}. FMCW radars are more common due to their low cost and simple architecture. FMCW radars operate in the frequency domain, translating time-of-flight into frequency shifts in their received signals. A Fast Fourier transform (FFT) is required to translate these signals back into the time domain and obtain range, velocity, and angle measurements. 

FMCW radars typically use a \textit{linear chirp}, i.e, a wave of linearly increasing frequency for a fixed duration, as the basic unit of the transmitted signal. The chirp waveform is reflected upon interacting with various agents. The resulting received signal is sampled and FFT-processed to obtain observations from objects at different ranges (known as echoes). The amplitudes of echoes at different range bins indicates the absence or presence of objects, along with their sizes and types. For example, large buses with metallic bodies generate large-amplitude echoes that span several range bins, while pedestrians generate small-amplitude echoes that span only a few bins. The geometric size and reflectivity of any object can be combined into a radar cross section (RCS). The latter can be used to analytically estimate the signal-to-noise ratio (SNR) of the object using the classical radar range equation~\cite{skolnik2001introduction}. Rao~\cite{SandeepRao} provides a great tutorial on working of FMCW radars. 

During a typical FMCW radar frame of about 100~ms, each transmitter emits multiple chirps. The FFT-processed data from these chirps are stacked to form a 2D matrix, on which an FFT is performed over the chirp dimension to calculate the radial velocity for each observation. Concurrently, data from multiple transmit-receive pairs is combined to generate a synthesized 2D array, as demonstrated in Fig.~\ref{fig:radar-basics}(b). As for the chirp dimension, an FFT is performed along the horizontal antenna dimension after stacking 2D FFT-processed matrices from horizontal antennas, thus resolving the azimuthal angle ($Az$) of the observations. Subsequently, 4D radars perform a similar FFT along the vertical antenna dimension to resolve the elevation angle of the observations. 

While the frequency domain architecture of FMCW radars renders them simple and affordable, it also leaves them vulnerable to substantial interference from neighboring FMCW radars due to frequency overlap. This problem has propelled development of coded radars, which transmit orthogonally-coded noise-like signals to minimize mutual interference. However, coded radars require high-speed signal processing due to their use of noise-like signals, which in turn makes them expensive and power-hungry and leaves FMCW radars as the more practical choice. Moreover, architecture-level strategies to mitigate interference in FMCW radars have been devised~\cite{Uysal2020PhaseCodedFA}, making them more usable in the presence of other FMCW radars. 

The first generation of FMCW radars employed in autonomous navigation were 3D devices that offered low resolution in azimuth, resulting in a relatively low detection probability. However, the increasing demand for high-resolution radars led to the development of next-generation 4D radars. These devices offer significantly improved resolution across all dimensions, including elevation, and thereby promise to  enhance the contributions of radar to autonomous perception.

\begin{table}[h]
    \centering
    \scriptsize
    \begin{tabularx}{\linewidth}{>{\raggedright}p{0.18\linewidth}
    >{\raggedright}p{0.10\linewidth}
    >{\raggedright}p{0.13\linewidth}
    >{\raggedright}p{0.13\linewidth}
    >{\raggedright}p{0.11\linewidth}
    >{\raggedright\arraybackslash}p{0.15\linewidth}}
      \toprule
      \textbf{Specification} & \textbf{4D Radar} & \textbf{3D Radar (Near)} & \textbf{3D Radar (Far)}& \textbf{32-Beam Lidar} & \textbf{Solid State Lidar}\\
      \midrule
      Max range & 300 m &  70 m & 250 m & 200 m & 200 m\\
      FoV (H/V) & 120$^o$/30$^o$ & 120$^o$/\ding{55} & 20$^o$/\ding{55} & 360$^o$/40$^o$ & 120$^o$/25$^o$\\ 
      Ang. res (H/V) & 1$^o$/1$^o$ & 4$^o$/\ding{55} & 1.5$^o$/\ding{55} & 0.1$^o$/0.3$^o$ & 0.2$^o$/0.2$^o$\\
      Doppler res. & 0.1 m/s & 0.1 m/s & 0.1 m/s & \ding{55} & \ding{55}\\   
      Point density & Medium & Low & Low & High & High\\
      Wavelength & 4 mm & 4 mm & 4 mm & 905 nm & 905 nm\\
      All-weather & \ding{51} & \ding{51} & \ding{51} & \ding{55} & \ding{55}\\
      Cost & Medium & Low & Low & High & Medium\\
      \bottomrule
    \end{tabularx}
    \caption{A comparison of capabilities of radars and lidars. Generally, 3D radars provide two types of low-resolution imaging -- near range and far range. 4D radars have substantially better imaging resolution and more capabilities than their 3D counterparts, thus bridging the performance gap with lidars. Here FoV refers to the ``Field of View'' while H/V denotes Horizontal/Vertical. }
    \label{tab:radar-vs-lidar}
\end{table}

\Figure[]()[width=0.95\linewidth]{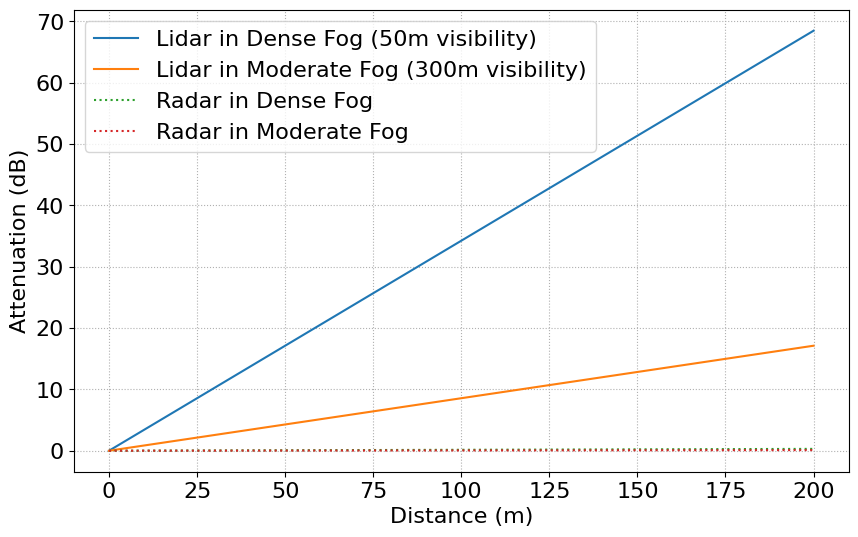}
{A comparison of signal power attenuation in moderate and dense fog conditions for 77 GHz Radar (using the ITU model) and 905 nm Lidar (using the Al Naboulsi model). This chart shows that lidar undergoes significant attenuation in fog while radar is unaffected. \label{fig:fog-attenuation}}

\Figure[]()[width=0.95\linewidth]{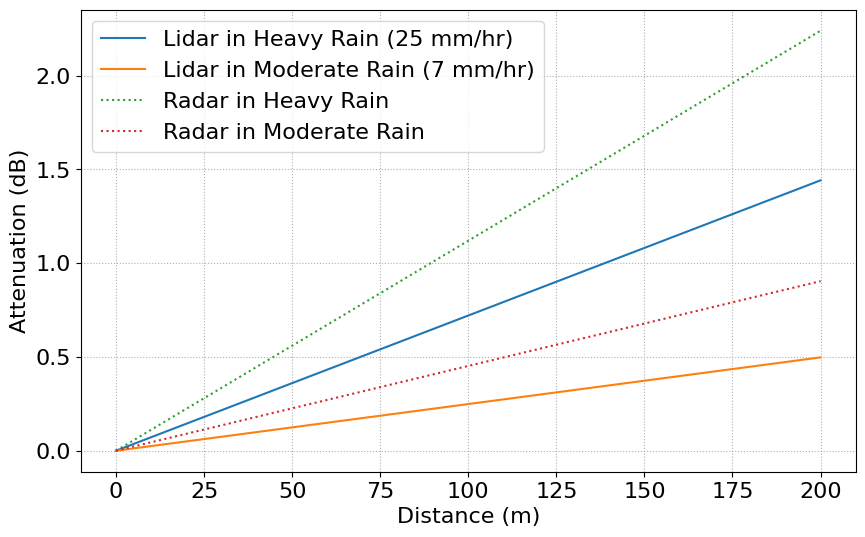}
{A comparison of signal power attenuation in moderate and heavy rain conditions for 77 GHz Radar (using the ITU model) and 905 nm Lidar (using a Marshall-Palmer rain distribution). This chart shows that both lidar and radar perform well, even in heavy rain conditions. \label{fig:rain-attenuation}}

\subsection{Comparison with lidar}
The new 4D radars represent a significant upgrade over their predecessors, offering 3D spatial and velocity imaging of objects. By contrast, lidars generate a dense, high-resolution 3D point cloud of objects over a 360$^{\circ}$ field of view. Traditional lidars, made of delicate spinning lasers, are cost-prohibitive and high-maintenance, thus they are primarily used in ride service-targeted autonomous vehicles and not in general-purpose electric vehicles. 

With the introduction of semiconductor chip-based solid-state lidars, a more affordable alternative to spinning lidars has become available. Despite having a lower resolution and a reduced field of view, these devices are better suited for general-purpose vehicles. In addition, a category of FMCW lidars has recently been developed that can detect the instantaneous radial velocity of objects, similar to radars~\cite{thinkautonomousUnderstandingMagnificent}. However, lidars still suffer from sensor degradation in adverse weather. Thus, a comparison of capabilities of lidars and radars in context of autonomous driving is pertinent. We present a detailed comparison of their specifications and their performance in adverse weather conditions in the subsections below. 

\subsubsection{Sensor Specifications}
For our comparison, we chose two generations of radars and lidars. For radars, we included the new-generation 4D radar and its predecessor, the 3D radar, which alternates between two modes (near and far) to achieve a balance of resolution and imaging range. For lidars, we selected the conventional 32-beam spinning lidar and the newer, cost-effective solid-state lidar. A side-by-side comparison of these sensors' specifications is presented in Table~\ref{tab:radar-vs-lidar}.

Upon analysis, we find that 4D radars have notably improved angular resolution and maximum range compared to 3D radars. By contrast, solid-state lidars are more affordable than spinning lidars but have reduced azimuth resolution and field of view (FoV). We also find an interesting similarity between the specifications of 4D radar and solid-state lidar: both sensors have similar cost and FoV. However, the solid-state lidar offers five-fold better angular resolution, resulting in denser point clouds. On the other hand, 4D radar excels in velocity imaging, offers a longer imaging range, and is resilient to weather conditions.

Thus, it is likely that autonomous vehicles will utilize both lidars and radars to ensure high levels of safety and reliability during full self-driving, while general-purpose electric vehicles will likely utilize radars, sometimes accompanied by solid-state lidars, to balance affordability and high levels of autonomy. We provide a more in-depth comparison of weather performance for the two sensors below.

\subsubsection{Adverse Weather Performance}
For the adverse weather performance analysis, we modeled the received signal attenuation for lidar and radar over a 200m distance in moderate and heavy rain (Fig.~\ref{fig:rain-attenuation}) and moderate and dense fog conditions (Fig.~\ref{fig:fog-attenuation}). Moderate and heavy rains are characterized by rain rates of 7 mm/hr and 25 mm/hr~\cite{baranidesignRainRate}, respectively. Similarly, moderate and dense fogs are characterized by visibilities of 300m and 50m~\cite{hindawiStudyLargeScale}, respectively.

\textbf{Radar signal attenuation modeling.} We modeled radar signal attenuation in rain and fog using the International Telecommunication Union's (ITU's) rain and fog models -- specifically, ITU-R P.838-3~\cite{ITU_Rain} for rain and ITU-R P.840-8~\cite{ITU_Fog} for fog -- which provide attenuation rates per kilometer in decibels (dB). It should be noted that these models are only applicable for frequencies up to 1~THz, making them unsuitable for lidar attenuation modeling.

\textbf{Lidar signal attenuation modeling.} The signal attenuation models in the visible spectrum estimate the extinction coefficient ($\alpha$), which can be employed to compute attenuation over a distance $d$ as:
\begin{equation}
    Attenuation(d) = e^{-\alpha d}
    \label{eq:attn}
\end{equation}
The attenuation in dB can then be further computed as $-10log_{10}(Attenuation(d))$.

For fog, the extinction coefficient can be accurately modeled using Al Naboulsi's fog models~\cite{Naboulsi2004FogAP}. In the case of advection fog, which is prevalent in areas like San Francisco, the formula is
\begin{equation}
    \alpha_{V,adv} = \frac{0.11478\lambda_{um} + 3.8367}{V} [m^{-1}],
    \label{eq:fog-exinction}
\end{equation}
where $\lambda_{um}$ represents the wavelength in micrometers and $V$ is the visibility in meters. The attenuation can then be calculated using eqn.~(\ref{eq:attn}).

Modeling the extinction coefficient for rain introduces greater complexity and is more susceptible to errors, given that the rain rate only indirectly informs about visibility and raindrop size. One method involves integrating over raindrop size distribution for a given rain rate $R$ to derive the rain extinction coefficient. The result is
\begin{equation}
    \alpha_{rain} = \int (q_{sca} + q_{abs}) N(r)\pi r^2dr
\end{equation}
where $r$ is raindrop size, $N(r) = N_0e^{-2\Lambda(R) r}$ is the Marshall-Palmer raindrop size distribution, and $q_{sca}$ and $q_{abs}$ are scattering and absorption extinction efficiencies for a given raindrop size as determined through Mie theory~\cite{oceanopticsbookTheoryOverview}. The attenuation can subsequently be calculated using eqn.~(\ref{eq:attn}).

Based on above models, Fig.~\ref{fig:fog-attenuation} compares the signal attenuation over distance for lidar and radar under foggy conditions. We observe that lidar is considerably impacted by fog while radar remains largely unaffected. This disparity arises because fog droplets are approximately the same size as the lidar wavelength, leading to substantial scattering of the lidar signal. This concludes that lidar is not much reliable in foggy settings.

Fig.~\ref{fig:rain-attenuation} provides a similar comparison for lidar and radar under rainy conditions. Compared to fog droplets, raindrops are notably larger (typically, on the order of millimeters) and are dispersed less densely. This reduces lidar signal scattering, resulting in a marked reduction in the attenuation of the lidar signal during rain. Intriguingly, our analysis reveals that radar experiences greater attenuation than lidar in rainy conditions, though the amount of signal loss is not prominent for either sensor. Instead, the primary challenge for these sensors in rainy conditions is the spurious observations (or points) that arise from reflections off raindrops. Such reflections can increase the number of false-positive object detections by learned models. This is a smaller problem for radar because most of these observations are nearly stationary and, thus, don't interfere with or lead to detections of moving objects.

\Figure[]()[page=8, width=0.99\linewidth]{figures/pcp-overview.pdf}
{A diagram showing different radar data formats obtained from the radar signal processing chain. The chain takes the sampled receive signal as input and is capable of generating a variety of formats, including the radar point cloud data, within the radar itself.\label{fig:radar-spc}}


\Figure[]()[page=6, width=\textwidth]{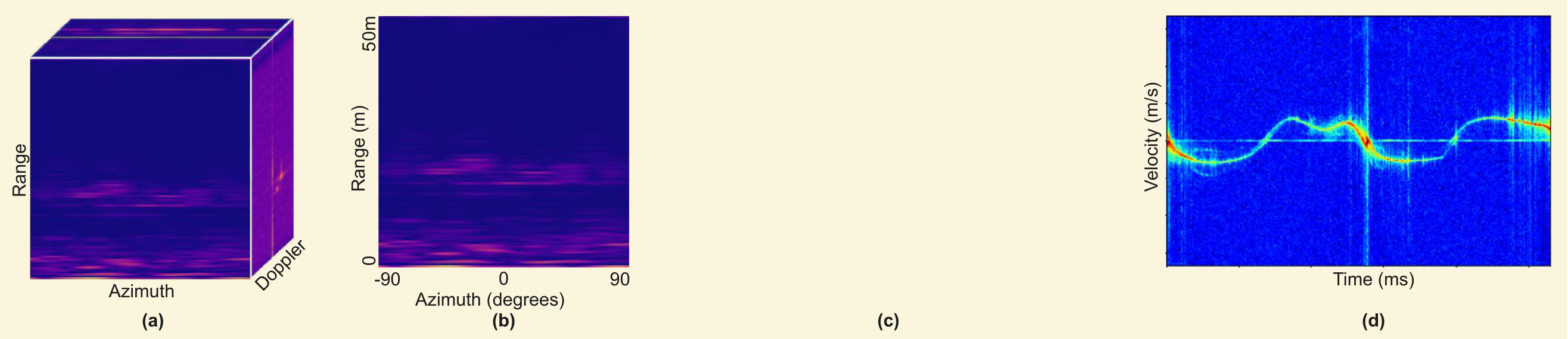}
{Illustration of commonly used radar data formats. (a) The range-azimuth-Doppler (RAD) tensor offers high resolutions along the range and Doppler dimensions but has a low resolution along the azimuth dimension. (b) The range-azimuth heatmap, a relatively coarser data format, is obtained by compressing the Doppler dimension of the RAD tensor. (c) The radar point cloud includes strong observations from the radarcube in a sparse point cloud representation. (d) The micro-Doppler spectrogram captures fine-grained temporal variations of object motion features, but does not provide spatial information.\label{fig:radar-representations}}


 \subsection{Radar data formats}
As previously discussed in Section~\ref{sec:radar_overview}.C, FMCW radars store the range, Doppler, and angle observations of agents for each frame in the multi-dimensional FFT-processed matrix, referred to as the \textit{radarcube}. For 3D radars, the size of a radarcube typically reaches $32$ millions cells ($1024$ range bins $\times 1024$ velocity bins $\times 32$ angle bins), and it is even larger for 4D radars. With a 16-bit value for each matrix cell, the data rate required to transfer radarcubes can reach 2.5~Gbps for 3D radars and 10~Gbps for 4D radars. Such large multi-dimensional radarcube sizes are impractical for deep learning as they imply the need for large convolutions at each layer of the model.
 
Fortunately, crucial observations within the radarcube are extremely sparse, as agents only occupy a few localized subspaces. Hence, it is common practice to extract useful observations in various compressed data formats for model development. These formats include the range-azimuth-Doppler (RAD) tensor, range-azimuth heatmap, point cloud, and micro-Doppler spectrogram, as illustrated in Fig.~\ref{fig:radar-representations}, with the point cloud being the most common format. These formats are extracted at various stages of the FFT-based signal processing chain in radars, as shown in Fig.~\ref{fig:radar-spc} We describe these formats below.

\subsubsection{RAD Tensor} 
The range-azimuth-Doppler (RAD) tensor is a 3D representation extracted from the radar signal processing chain following the azimuth angle FFT. Typically, the observation values are averaged over the elevation dimension, if present. In addition, the tensor is occasionally downsampled across the three dimensions to reduce its size. Despite being relatively large, the RAD tensor offers the benefit of localizing agents in the top-down view (via range-azimuth), while retaining robust radar features within the high-resolution range and Doppler dimensions. Consequently, this format allows learned detection models to infer object attributes, such as position, speed, and 2D shape, directly from the tensor.

\subsubsection{Range-Azimuth Heatmap}
The range-azimuth heatmap is a 2D radar image (top-down view) obtained by compressing the Doppler dimension of the RAD tensor. Each bin in the heatmap carries the amplitude of the observations, occasionally including averaged Doppler. While the heatmap provides a more manageable data size for deep learning models compared to RAD tensors, it loses the rich radar features essential for high performance of learned detection models. This is due to the compression of fine-grained Doppler features and the high uncertainty associated with low azimuth resolution. 

\subsubsection{Radar Point Cloud}
The radar point cloud is a commonly used radar data format that represents the relevant observations within the radarcube as a sparse set of data points in a 3D space. Each point in the cloud corresponds to a peak in the radarcube, obtained using a constant false alarm rate (CFAR) peak extraction algorithm. CFAR classifies a cell in the radarcube as containing a signal (i.e., echo) if its amplitude relative to the average noise floor (computed from surrounding cells) exceeds a certain threshold. The value of this threshold can be adjusted to set the desired false alarm rate, i.e., incorrect classification of noise as a signal of interest. Each point in the cloud contains both 3D location features consisting of range, azimuth, and elevation (if present), and also radar-specific features that include radial velocity, SNR, and measured RCS.

The point cloud format is especially beneficial for object detection and classification models, as it substantially reduces data dimensionality while preserving the essential details of the object and the scene. However, the point cloud format may filter out weaker (i.e., low SNR) observations from agents, which may be undesirable for a high-performance radar deep learning model.

find and cite the paper on improving learning from noise
\Figure[]()[width=0.99\linewidth]{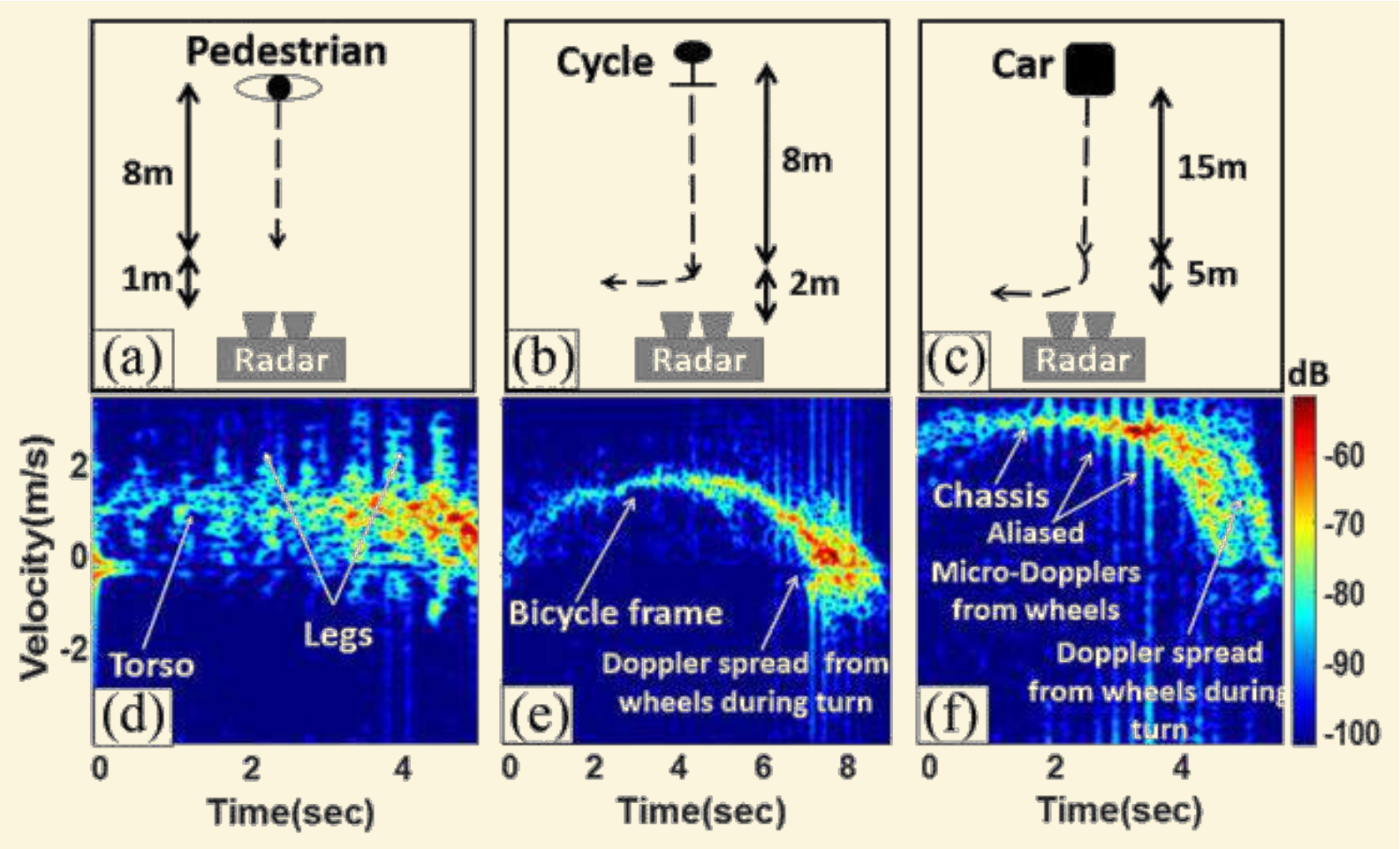}
{The micro-Doppler spectrograms of a pedestrian (d), a bicycle (e), and a car (f) for their motions outlined in (a-c)~\cite{Duggal2019DopplerResilient8U}. These spectrograms capture distinct motion patterns of different parts of these agents, as illustrated in these plots. During turning, the velocity features of the agents become more complex as their body parts undergo different rapidly changing motions.\label{fig:micro-Doppler-spectrogram}}


\subsubsection{Micro-Doppler Spectrogram}
The micro-Doppler spectrogram is a 2D representation that maps the Doppler frequency shift against time~\cite{microDopplerdeWit}. It is a classical format obtained by performing a short-time Fourier transform after the range FFT in the radar signal processing chain. This format excels in capturing rich motion features of agents. These features are quite distinctive, both for different agents and their different motion types, as illustrated by the micro-Doppler features from moving and turning events of a car, a bicycle, and a pedestrian shown in Fig.~\ref{fig:micro-Doppler-spectrogram}. Patterns within the micro-Doppler spectrogram can be used to classify the type of object and infer its activities. An interesting application of this format to detect drones and birds was studied in~\cite{Rahman2018RadarMS}.

The 2D spectrogram format does not capture spatial features, which makes it unsuitable for object detection models. However, this format underscores the importance of time-varying Doppler features in radar data for reliable object detection. 

\Figure[]()[page=5, width=0.99\linewidth]{figures/radar-data.pdf}
{Different radar data input formats used for deep learning models. The bird-eye view (BEV) (a) represents data in a top-down grid, while the perspective view (b) projects data as seen from an image perspective.\label{fig:radar-data-formats}}


\begin{table*}
\centering
\caption{A comparison of recent high-quality radar datasets for autonomous systems}
\label{tab:datasets}
\begin{tabularx}{\textwidth}{P{1.55cm} P{1.45cm} P{0.75cm} P{0.75cm} P{3cm} P{2cm} P{1.25cm} P{2cm} P{1.25cm}}
    \toprule
    \textbf{Dataset} & \textbf{Tasks} & \textbf{Citations} & \textbf{Year} & \textbf{Radar Data} & \textbf{Other Sensors} & \textbf{Data Size} & \textbf{Diversity} & \textbf{Annotations}\\
    \midrule
    K-Radar \cite{Paek2022KRadar4R} & Detection, Tracking & 3 & 2022 & 4D Radar Tensors and Point Cloud from Front Radar  & Lidars, 4 Cameras & 35K Frames & Different Weather and Lighting, Korea & 93K 3D Bboxes\\
    \midrule
    View-of-Delft \cite{Palffy2022MulticlassRU} & Detection, Tracking & 22 & 2022 & 4D Radar Point Cloud from Front Radar & 64-Beam Lidar, Stereo Camera & 8.6K Frames & Sunny, Dense Traffic, Netherlands & 123K 3D Bboxes\\
    \midrule
    NuScenes \cite{Caesar2019nuScenesAM} & Detection, Tracking & 2877 & 2019 & 3D Radar Point Cloud from 5 Surround Radars & 32-Beam Lidar, 6 Cameras & 1.3M Frames & Urban Scenes, Boston, Singapore & 1.4M 3D Bboxes\\
    \midrule
    RadarScenes \cite{Schumann2021RadarScenesAR} & Detection, Tracking & 52 & 2020 & 3D Radar Point Cloud from 4 Surround Radars & Cameras & 118M Points & Urban Scenes, Germany & Point-wise\\
    \midrule
    Boreas \cite{burnett2023boreas} & Detection, Tracking, Localization & 24 & 2022 & Range-Azimuth Heatmap from Top Rotating Radar (No Velocity) & 128-Beam Lidar, Front Camera & 7.1K Frames & Sunny, Toronto, Canada & 320K 3D Boxes\\
    \midrule
    Oxford Radar Robotcar \cite{RadarRobotCarDatasetICRA2020} & Localization & 249 & 2020 & Range-Azimuth Heatmap from Top Rotating Radar (No Velocity) & 2 32-Beam Lidars, Front Camera & 240K Frames & Varied Lighting, Limited Weather, UK & N/A\\
    \bottomrule
\end{tabularx}
\end{table*}

\subsection{Input formats for deep learning models}
While the formats mentioned above are commonly used to represent radar data, object detection and perception models often use different representations for input data. Typically, these models detect and track objects in either a bird-eye view or a perspective view. We discuss these representations for the radar point cloud below.

\subsubsection{Bird-Eye View (BEV)}
The bird-eye view (BEV) offers a top-down view of the scene data. This representation excels in preserving distances between agents in the scene and separating all objects spatially, including those in occlusion. The conversion of radar point cloud data to BEV usually involves accumulating points within each cell of a top-down grid, followed by learning fixed-sized features within each cell to retain most of the properties present in the point cloud data~\cite{Lang2018PointPillarsFE}.

The BEV projection compresses the spatial dimensions of the data from 3D to 2D, thus reducing the required complexity of the detection models. However, the size of the 2D grid expands quadratically with increase in either the range or the grid resolution of the BEV representation, making the detection models quadratically larger. As a result, this representation is not ideal for long-range detection models. 

\subsubsection{Perspective View}
The perspective view is the view normally seen by human eyes. It is similar to camera images from the ego vehicle, but with an added depth feature (and accompanying velocity feature for radar data) assigned to each pixel. Unlike the BEV, the size of the grid in this view remains constant, regardless of range. This characteristic makes it a suitable choice for long-range detection tasks. Fig.~\ref{fig:radar-representations}(d) shows the perspective view of a radar point cloud.

Despite this advantages, using a perspective view for radar data presents considerable challenges. At each pixel behind an observation, radars often register several weaker observations from longer ranges that can be obscured in the perspective view. Moreover, radars provide low-resolution perspective view images due to their low azimuth and elevation resolutions, which further diminishes the effectiveness of this representation.


\section{Autonomous radar datasets}
\label{sec:datasets}
High-quality datasets are the backbone of cutting-edge deep learning research. The proliferation of computer vision research, for instance, was largely propelled by the availability of the extensive and high-quality ImageNet dataset~\cite{ILSVRC15}. Similarly, for autonomous vision and lidar research, high-grade, large-scale datasets like the KITTI robotics dataset~\cite{Geiger2013IJRR} have been available since 2013. Over the years, the quality of these datasets has continued to improve, with offerings such as NuScenes~\cite{Caesar2019nuScenesAM} and the Waymo Open Dataset~\cite{Sun2019ScalabilityIP}.

However, the evolution of radar datasets has not been as smooth. The first radar datasets for autonomous applications, the multimodal NuScenes dataset~\cite{Caesar2019nuScenesAM}, DENSE\cite{Bijelic2019SeeingTF} and RADIATE~\cite{Sheeny2020RADIATEAR}, were released around 2019. But the quality of radar data in these datasets was somewhat lacking, primarily due to the limitations of 3D radars. Additionally, these datasets contained a limited amount of labeled radar data. Fortunately, new and more promising autonomous radar datasets are surfacing with the development of next-generation 4D radars, as required to advance radar deep learning research.

The most useful publicly-available radar datasets are summarized in Table~\ref{tab:datasets}. For each dataset, we have noted their targeted tasks, popularity gauged by citation count, release time, a description of the radar data, additional sensor data included, the size of the dataset, its diversity, and the size and type of annotations. This comprehensive summary should serve as a valuable resource when selecting the right dataset for model development. The list is arranged in order of decreasing assessed value.

Next, we provide a brief overview of the datasets. K-radar~\cite{Paek2022KRadar4R}, though recent, is a high-quality radar dataset. It includes dense 4D radar tensors along with 4D radar point cloud data, contains a wide diversity of scenes, and has a large size. However, it only contains data from front-facing radar. The View-of-Delft (VoD) dataset~\cite{Palffy2022MulticlassRU} also offers high quality 4D radar point cloud data from front-facing radar, but falls short in terms of scene diversity and size compared to K-radar. 

On the other hand, the most popular radar dataset for research at present is the multimodal NuScenes dataset, which is larger in size compared to K-radar but contains radar point cloud data from previous-generation 3D radars. It has also been extensively used to develop radar-based early fusion models, discussed later. RadarScenes~\cite{Schumann2021RadarScenesAR} is another large radar dataset worth considering, particularly for models requiring point-wise annotations. Finally, Boreas~\cite{burnett2023boreas} and Oxford Radar Robotcar~\cite{RadarRobotCarDatasetICRA2020} offer radar range-azimuth heatmap data, albeit without velocity information, and they have been used for a few object localization and lidar-radar early fusion model studies~\cite{Qian2021RobustMV}.
 

\Figure[]()[page=5, width=\textwidth]{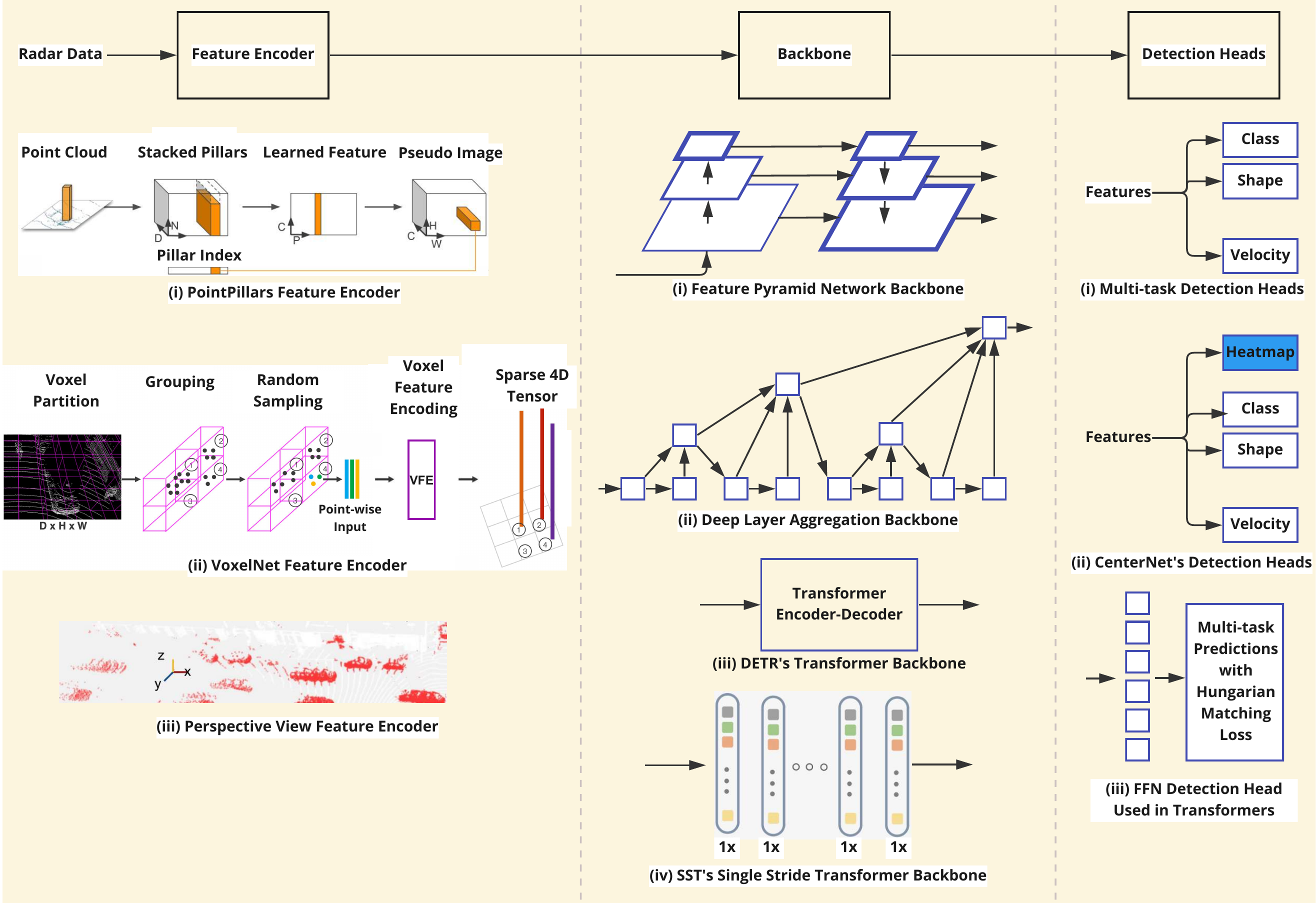}
{These high-level block diagrams represent popular paradigms in autonomous perception. Diagram (a) depicts late fusion detection-based tracking, which fuses predictions of individual object detection models. Diagram (b) represents early fusion detection-based tracking, which jointly detects objects via early fusion models and follows this by association and tracking. Diagram (c) illustrates occupancy-based tracking, which predicts fine-grained scene occupancy for estimating the drivable surface area, as well as for object detection and occupancy tracking. \label{fig:pcp-overview}}


\section{High-level overview of perception}
\label{sec:pcp-overview}
At a high level, autonomous perception typically has two steps: detection and tracking. Detection involves identifying objects in the scene from sensor data. For example, identifying other vehicles, bicyclists, pedestrians, and road signs in the images or point clouds captured by the sensors. Tracking, on the other hand, involves following the identified objects over time~\cite{pendleton2017perception, Bewley2016SimpleOA, Voigtlaender2019MOTSMT}. 

Traditional methods have used a late fusion approach for detection in which results from individual sensor detection models are fused, but more recent methods have begun exploring early fusion where raw features from multiple sensors are fused in a model to infer joint detections~\cite{thinkautonomousLiDARCamera}. Additionally, there are methods that eschew detection altogether for tracking, such as occupancy-based methods. We provide an overview of late fusion, early fusion, and occupancy-based methods below. 

\subsection{Detection-based tracking}
\subsubsection{Late Fusion}
The late fusion approach employs separate full object detection models on each sensor's data to infer the class, distance, shape, and orientation attributes of agents in the scene. Some dissimilarities in the attributes from different sensor models exist, such as cameras providing 2D perspective view detections, lidars 3D BEV detections, and radars 2D BEV detections with associated velocities. 
 
The detections from different models are associated together to infer unified perception detections, commonly using a learned associator~\cite{Zhang2021ByteTrackMT}. The associator produces the unified detections in a joint 3D space and learns to assign weights to model outputs from sensors based on their respective strengths. For instance, in the unified output, it might assign greater weights to class, shape, and orientation attributes from lidar and camera detections and to velocity attributes from radar detections. The tracker updates agent tracks in the current frame by using tracks from the preceding frame and detections in the present frame, commonly by applying classical models such as the extended Kalman filter~\cite{kalman_filter} or learned models like the unsupervised learned Kalman filter~\cite{revach2021unsupervised}.

In the early days of autonomous driving, the late fusion approach was widely adopted due to its ease of implementation and iteration. It was challenging to develop unified detection models on disparate data from different sensors. By contrast, the detection models for each sensor produced similar outputs that could be effectively combined using an associator. This approach also facilitated quick upgrades of individual sensor models as more advanced detection models emerged from research.  

While the late fusion approach is generally effective, it can struggle in reliably detecting agents with weak or partial signals in multiple sensors. Large vehicles such as vans or trucks positioned directly ahead of the ego vehicle are a good example. 
Lidars capture limited points from the side surfaces of these vehicles when they are directly ahead. By contrast, radars yield dense uniform returns from these agents, owing to their large metallic bodies and underbody reflections~\cite{Kopp2021FastRC}, as discussed in Section~\ref{sec:challenges}. Nevertheless, the associator in a late fusion system may routinely disregard shape features from radar model outputs due to the absence of in-context learning. A similar situation arises when driving in adverse weather, where the associator might continue to underweight the shape features of radar model outputs, even though these features might be more reliable than those from the lidar model output. These challenges of late fusion have led to a growing interest in early fusion, which is discussed below.
 
\subsubsection{Early Fusion}
One of the main limitations of the late fusion approach is its association of model outputs that have stripped away all data features. By contrast, early fusion models learn object detections jointly from multiple sensor inputs, which give them access to data-level features from the sensors. These models effectively perform the roles of both multiple single-sensor models and the associator within the late fusion pipeline, but in an improved fashion. Early fusion has gained further traction with the advent of transformers, which enable more effective joint learning from dissimilar multi-modal data through their cross-attention mechanism. In fact, recent early fusion models have demonstrated notable improvements in detection performance compared to their single-sensor counterparts~\cite{bai2022transfusion, Qian2021RobustMV}. 

Thus far, the bulk of early fusion research has focused on the fusion of data from two sensors such as camera-radar, lidar-radar, and camera-lidar. Thus, an associator remains necessary to produce final unified detections for tracking. Fig.~\ref{fig:pcp-overview} presents a high-level overview of both late fusion and early fusion methods in perception, along with an emerging occupancy-based tracking method which we describe below.

\subsection{Occupancy-based tracking}
A significant drawback of the detection-based tracking paradigm for perception is its dependence on object detection, which can be unreliable for rarely observed agents such as rolling tires, reflective trucks, protruding construction vehicles, or workers emerging from manholes, among others~\cite{datagenEdgeCases}. Given the limitless variety of such uncommon scenarios, there will inevitably be elements that are under-represented in model datasets. 
 
However, object detection is not essential for tracking or ego vehicle navigation. Fundamental principles of robotic autonomy suggest that a safe navigation of the ego vehicle is feasible if the occupancy and velocity of cells in a fine-grained 3D gridded world around the ego vehicle are known. Occupancy-based models aim to do just this by developing an ego-centric, fine-grained 3D grid of the world and determining the occupancy and velocity of each cell. Such a representation allows for the estimation of occluded regions, drivable areas, and moving agents, which in turn facilitates tracking, prediction, and planning for ego vehicle navigation. 

Occupancy-based approaches are effective in handling rare objects since they do not rely on object detection priors for tracking. All objects occupy cells that can be tracked and constitute non-drivable regions regardless of their rarity or under-representation in data. Another strength of these methods is their reduced dependency on labeled data for model training as occupancy estimation can be learned from sensor data using semi-supervised methods.

Occupancy estimation is more reliable at short-to-medium ranges than at long ranges due to the increasing uncertainty of sensor data with range. In both occupancy-based and detection-based methods, a significant source of uncertainty is sensor noise in dynamically changing environments, known as \textit{aleatoric heteroscedastic uncertainty}. The regions with high aleoteric heteroscedastic uncertainty in sensor data typically yield low confidence occupancy predictions. Recognizing and harnessing this uncertainty can enhance the robustness of the occupancy models. We delve deeper into this topic in Section~\ref{sec:uncertainty}.

\Figure[]()[width=\textwidth]{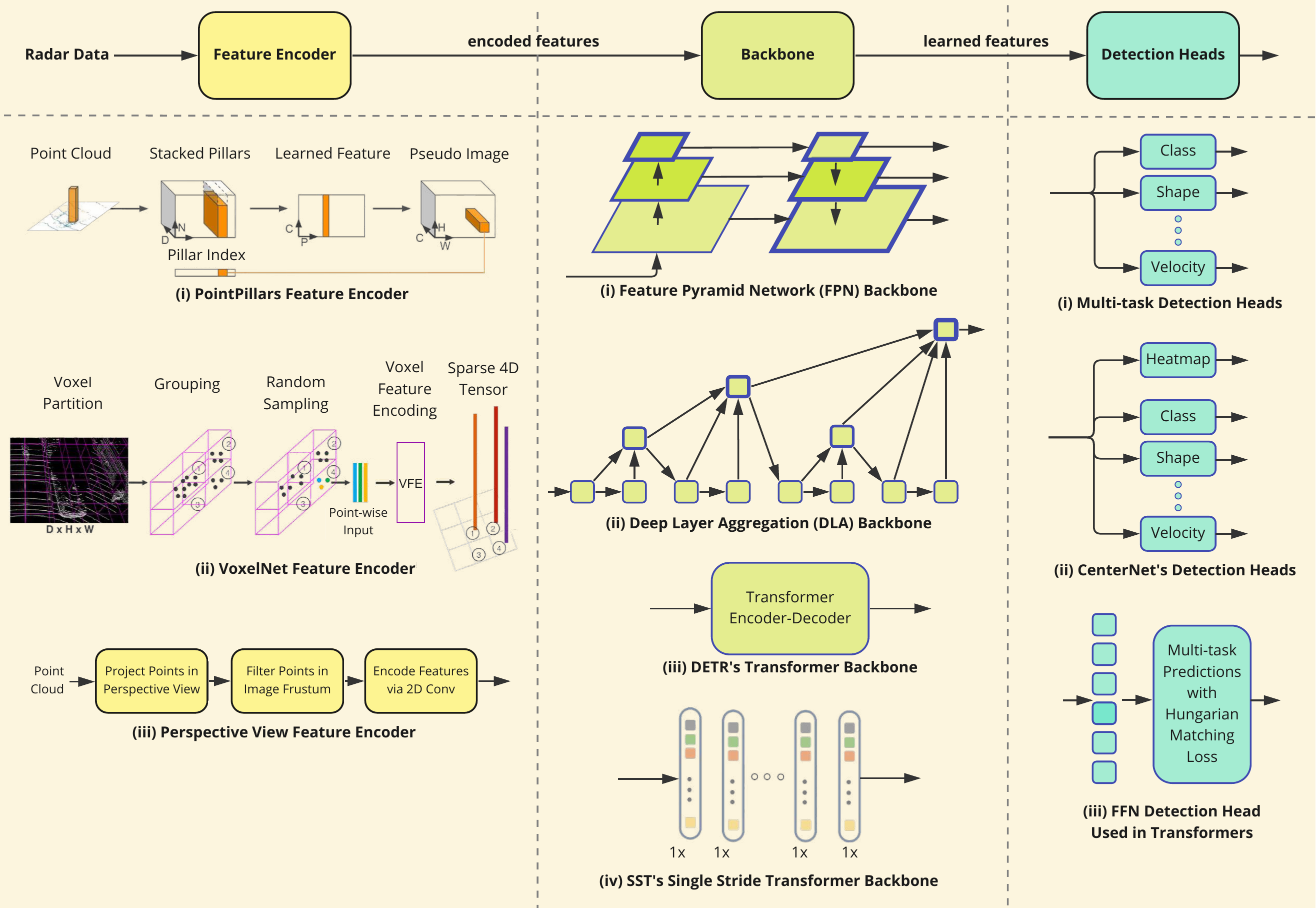}
{A high-level architectural block diagram of radar models, which typically consist of three stages: a feature encoder, a backbone, and detection heads. The feature encoder processes radar data, often in point cloud format, and generates encoded BEV or perspective view features. The backbone, comprising of several convolutional or transformer-based layers, takes in the encoded features and outputs learned features. Detection heads take in the learned features and predict various detection attributes using multiple heads in convolutional models or a feed-forward network (FFN) head in transformer-based models. Notable components from leading studies are shown for each stage. \label{fig:radar-only-ml}}



\section{Radars in detection-based tracking}
\label{sec:detection-based-tracking}

\subsection{Radar detection models}
Radar detection models for perception typically provide a comprehensive set of attributes such as class, distance, shape, orientation, and velocity for each detection via multi-task learning~\cite{Zhu2019ClassbalancedGA}. These attributes are utilized by the late fusion associator to generate unified object detections. Of these attributes, velocity and distance are the primary contributions from radar models to the unified late fusion output. Additional minor contributors include robustness to weather conditions and early detection of dynamic agents in occlusions or over long ranges. 

As previously noted, the majority of radar models are influenced by vision and lidar detection models due to challenges associated with the radar data and the scarcity of radar datasets. A handful of studies propose radar-based models, such as \cite{radarPointSemseg}, which performs semantic segmentation of radar point cloud, and \cite{Sun2022R2PAD}, which detects objects from the RAD tensor. These studies, however, utilize small, custom datasets and have a relatively narrow scope.

A high-level architectural block diagram of radar models is shown in Fig.~\ref{fig:radar-only-ml}. The input radar data, usually in point cloud format, is encoded using feature encoders in one of the popular perception input representations formats, as discussed in Section~\ref{sec:radar_overview}.F. These encoded features are then used by a backbone to extract learned features at varying levels, which are subsequently utilized by detection heads for multi-task predictions.

This design follows a modular detector framework, in which existing components can be replaced with components from various state-of-the-art models at relevant stages with minor modifications. We examine such state-of-the-art models and illustrate components from them at the appropriate stages in the architectural block diagram. We begin with a review of feature encoders, followed by an examination of backbones and detection heads for radar point cloud data.

\textbf{Feature encoders.} Point clouds are an unstructured data format and therefore unsuitable for tensor-based convolutional models. PointPillars~\cite{Lang2018PointPillarsFE, Shi2022PillarNetRA} is a pioneering work that proposed a pillars-based feature encoder to transform point cloud data into a structured BEV pseudo-image representation. It achieves this by employing stacked vertical pillars in a BEV 2D grid, aggregating points that fall within each pillar, and subsequently generating fixed-size learned features for each of these pillars. This results in a 2D pseudo-image, which is used by a 2D convolutional backbone to extract features at various levels. These features are then fed into a single-shot detection head~\cite{Liu2015SSDSS} for prediction.

Although the feature encoder proposed in PointPillars is lightweight and effective, its pillars lack height differentiation, leading to a significant loss of vertical structure encoded in the elevation of points. VoxelNet~\cite{VoxelNet, wang2019voxelfpn} is an extension of the PointPillars architecture to 3D which preserves the vertical structure of the data. Its voxel feature encoder converts raw point cloud data into a volumetric representation, or ``voxel grid'', using several fully connected layers to learn fixed-sized features for each voxel. This process results in a 4D tensor, which is then passed through a convolutional backbone followed by a region proposal network~\cite{Ren2015FasterRT} to generate 3D bounding boxes.

VoxelNet achieves enhanced detection performance over PointPillars, particularly for high-resolution and dense point clouds. However, it has a higher memory and computational footprint than PointPillars due to its 3D voxel grid. Thus, PointPillars is often a better choice than VoxelNet for radar data due to its lower density and resolution of points.

PointPillars and VoxelNet are ideal for short and medium-range BEV detections, but not for long-range detections due to the quadratic model complexity associated with range scaling. To overcome this, Range Sparse Net~\cite{sun2021rsn, Tian2022FullyCO} projects points in the perspective view and encodes point features in the perspective view grid. It subsequently uses an image-space backbone to extract multi-level features followed by detection heads to generate predictions.

\textbf{Backbones and detection heads.} There are two popular types of backbones in the literature: convolutional and transformer-based, each with its own type of detection heads. Convolutional backbones usually extract multi-scale features by down-sampling successive convolutional layers from their input, after which several deconvolution-based detection heads are implemented for multi-task object detection. By contrast, transformer-based backbones utilize their self-attention mechanism to learn object-level features, which are then used by feed-forward network (FFN) based detectors for inference. Though the transformer-based models generally perform better, their gain comes at cost of larger model sizes that require more training data. 

Initially, convolutional backbones commonly employed a detection head only on the final backbone layer output for predictions. However, this approach was less effective, as the resulting bottleneck failed to use features from previous layers. The feature pyramid network (FPN)~\cite{lin2017feature} addressed this problem by utilizing features from all layers of a backbone. By implementing feature pyramids via top-down and lateral connections in the deconvolution layers and attaching detection heads to all deconvolution layers, an FPN obtains detections at different feature scales. Subsequently, it combines detections via non-max suppression (NMS), which improves accuracy while keeping latency in check. Deep layer aggregation~\cite{yu2019deep} further improved upon FPN by using hierarchical and iterative skip connections from previous layers that deepen the representation and refine its resolution.

The backbones described above have state-of-the-art object detection performance on perspective view image datasets. However, in a BEV autonomous driving scene, objects exhibit significantly less variation in scale and are small relative to the image size. It has been argued that multi-stride backbones for BEV object detection bring little advantage and in fact lead to inevitable information loss due to the coarser resolution of the downsampled layers~\cite{fan2021embracing}. Through an experiment using the PointPillars model, the authors demonstrate that single-stride backbones with an appropriately-sized receptive field and without down-sampling provide better detection performance than multi-stride backbones. They further argue that transformer backbones are superior to convolutional backbones for BEV detections as they are better at learning features for objects that are small relative to image size. Building on these insights, the authors propose a single-stride sparse transformer (SST) architecture that takes voxelized input and implements a backbone consisting of single-stride SWIN transformer blocks~\cite{liu2021swin} to generate a dense feature map for feeding to the detection heads. SST achieves a substantial improvement in detection performance over state-of-the-art multi-stride backbone-based models on the Waymo Open Dataset.     

DETR~\cite{Carion2020EndtoEndOD} proposes a transformer encoder-decoder backbone for feature extraction followed by an FFN with bipartite Hungarian matching loss~\cite{Kuhn1955TheHM} for end-to-end object detection. DETR is simple in design and fully differentiable as it does not use any non-differentiable components, such as NMS, that are commonly used in convolutional architectures. 

Masked autoencoders (MAE)~\cite{He2021MaskedAA} provide a self-supervised pre-trained backbone for developing fully trained models with a small labeled dataset. During pre-training, a MAE segments the input into a top-down 2D grid (patches) and masks over $75\%$ of randomly selected patches. From the unmasked patches, a transformer encoder generates encoded features, which, along with mask tokens, are fed to a simple decoder to reconstruct the input. Afterwards, a detector model consisting of the pre-trained backbone and detection heads can be fine-tuned using a small labeled dataset.

The models discussed so far provide full object attributes predictions, including class, shape, and orientation. However, these attributes are learned from optical features of the data, which are relatively weak in radar data, leading to reduced detection performance on radar data processed by such models. 

CenterNet~\cite{Zhou2019ObjectsAP, Yin2020Centerbased3O} is a notable work that eliminates the need to predict full object attributes upfront. Instead, it models objects as their center points and trains the model parameters to estimate centers as keypoints. The remaining attributes of objects, such as shape, position, and orientation, are regressed subsequently. Hence, CenterNet presents an arguably better-suited architecture for radar-based object detection. CenterNet also proves useful as a detection head for perspective radar detectors, as the object attributes in radar data are even more noisy when represented in a low-resolution azimuth-elevation space~\cite{Nabati2020CenterFusionCR}.   

 \Figure[]()[page=6, width=\textwidth]{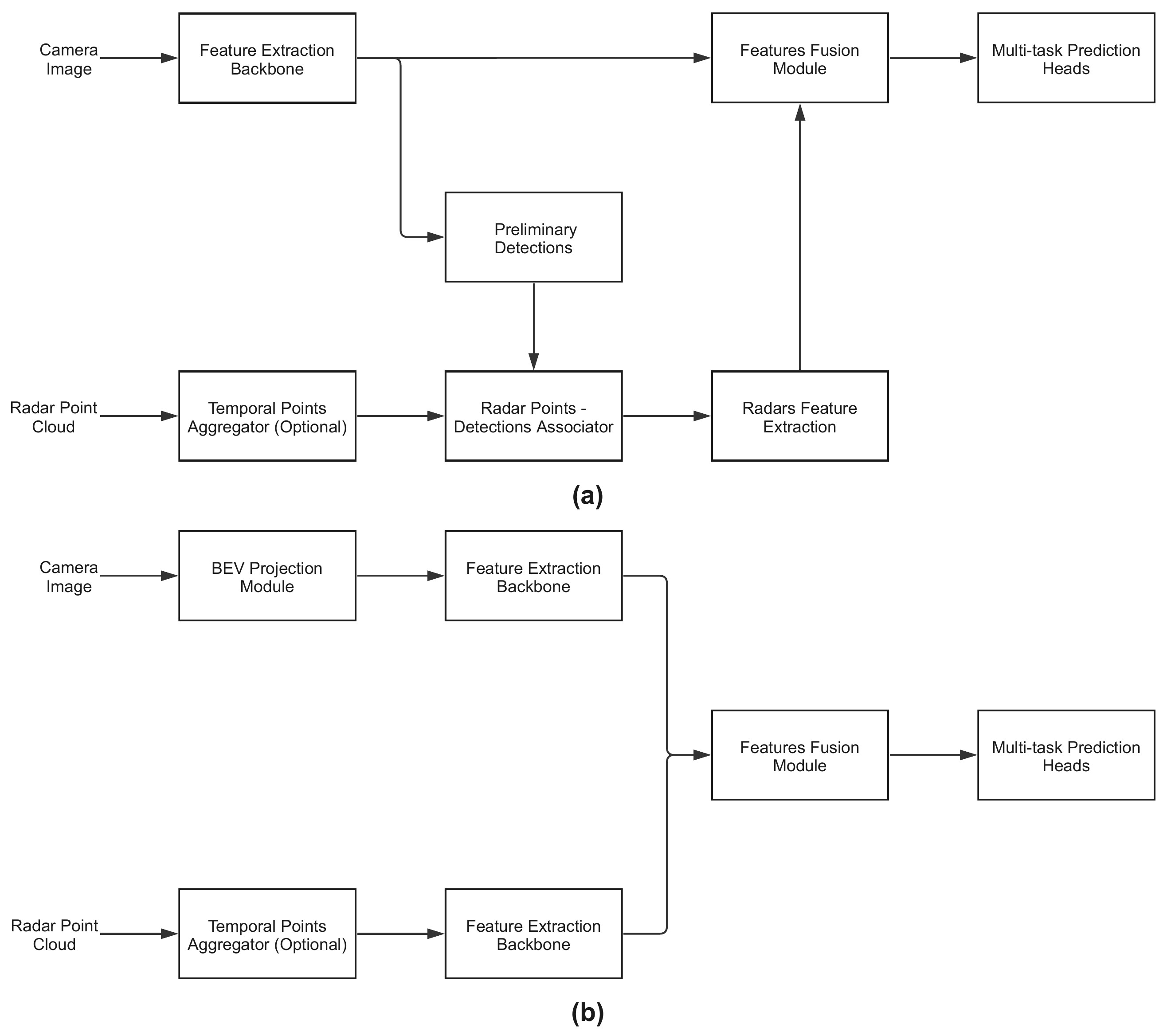}
{These high-level block diagrams depict common camera-radar early fusion models. Diagram (a) illustrates fusion in the perspective view, where radar points are associated with preliminary vision detections to extract learned radar features. The features are then fused with camera features to improve depth and velocity attributes of the preliminary detections. Diagram (b) shows fusion in the BEV, in which camera images are projected in the BEV by a learned model. The extracted BEV features of camera and radar are then fused to infer joint detections. \label{fig:cr-fusion}}

 
\subsection{Early fusion with radar}
Radar-based early fusion models learn from features of both radar and optical sensor data. Such joint learning results in improved detections, with optical data features contributing accurate optical attributes and radar data features complementing them with the unique strengths of radar. The two most prevalent fusion pairings with radar data are camera-radar fusion and lidar-radar fusion. In both cases, radar serves a complementary role, enriching detections with its strengths such as velocity, distance, and robustness to weather and lighting conditions. Initially, the majority of early fusion models in literature were convolution-based, though transformer-based models have recently gained popularity. These models demonstrate effective joint learning through their cross-attention mechanism, albeit at the cost of increased model size.
 
\subsubsection{Camera-Radar Early Fusion}
Camera and radar are both cost-effective and low-maintenance sensors with excellent complementary capabilities. Cameras capture detailed semantic information of the scene in a perspective view up to long distances, while radars provide accurate depth and velocity imaging, detections in occlusion and over long ranges, and robustness to weather and lighting conditions. Combined, they offer rich complementary features to facilitate advanced levels of autonomy.

However, early fusion of camera and radar presents a significant challenge due to their view disparity. Cameras image the scene in the perspective view, while radars capture rich features in the BEV. Converting one sensor's data to another's view is a complex and error-prone task. Although models such as Lift-Splat-Shoot~\cite{Philion2020LiftSS} can convert perspective view images to BEV, the results are not always accurate due to the ill-posed nature of camera monocular depth estimations. Similarly, representing radar data in the perspective view increases its ambiguity due to low azimuth and elevation resolutions of radars. 

The camera-radar early fusion models in the literature primarily use either a perspective view~\cite{Nabati2020CenterFusionCR, Long2022RADIANTRA} or a BEV~\cite{Kim2023CRNCR, Pang2023TransCARTC}, depending on their application. Some models, however, use non-standard joint views that offer a balanced compromise in view disparity for both sensors, such as CramNet~\cite{Hwang2022CramNetCF}.

CenterFusion~\cite{Nabati2020CenterFusionCR} is a convolutional early fusion model that uses perspective views to perform fusion of learned features from camera and radar to improve depth and velocity estimations of camera 3D detections. It employs a CenterNet-based architecture with the DLA backbone for extracting camera features and predicting preliminary 3D detections. Afterwards, it filters radar points falling in the frustum of each camera detection and concatenates learned features from those points with camera features for inferring final 3D detections. However, this model has issues with associating radar points to camera detections, especially when points at different depths fall within the same frustum. This may result in imprecise depth estimations.

CramNet~\cite{Hwang2022CramNetCF} fuses learned features from perspective view camera and BEV radar data in a ray-constrained projected 3D space via a transformer-based cross-attention mechanism. 
It also implements sensor dropout training by feeding Gaussian noise to one sensor input at random and demonstrates some model robustness to sensor data corruption in situations like adverse weather. However, it faces limitations in fusing features in the projected 3D space due to the lack of elevation resolution in the radar data. 
 
RADIANT~\cite{Long2022RADIANTRA} addresses the issue of imprecise association of radar points with camera object detections, which can lead to sub-optimal depth estimations. Instead of associating radar points to objects as other models do, RADIANT learns to predict 3D offsets between radar points and object centers, followed by a feedforward depth weighing network to refine their monocular estimated depth. This method could be extended for learning the velocity of objects as well. 

Contrary to the previous studies, Camera Radar Net (CRN)~\cite{Kim2023CRNCR} performs fusion in the BEV, using a multi-model deformable cross-attention module to fuse multi-view image and 3D radar point cloud data. It introduces some innovative techniques to mitigate issues with camera-to-BEV projection and BEV range scaling. It uses a radar-assisted view transform module to project image features in BEV that outperforms Lift-Splat-Shoot~\cite{Philion2020LiftSS} by including depth signals from radar data. In addition, its use of sparse aggregation by querying select features makes it efficient for long-range detection. These key innovations result in a $\sim$100~m range for the model and comparable detection performance to state-of-the-art lidar detector models on the NuScenes dataset. 

TransCAR~\cite{Pang2023TransCARTC} is another BEV early fusion model that utilizes DETR3D~\cite{Wang2021DETR3D3O} to generate 3D object features from multi-view images and uses an FFN on radar point cloud data followed by positional encoding to generate spatial radar features. Subsequently, it fuses camera and radar features via three multi-head cross-attention decoders, assisted by queries to a features closeness mask. The model achieves excellent results, but is large and slow due to its heavy reliance on transformers. 

While the models discussed above feature innovative architectures for camera-radar early fusion, they still face challenges due to issues with 3D radar point cloud data and disparities between camera and radar views. Apart from such models, an interesting study estimates the full instantaneous velocity of objects from their camera optical flow and radar radial velocity features using a simple closed-form solution~\cite{Long2021FullVelocityRR}. The authors demonstrate the accuracy of their velocity estimation by precise accumulation of radar points from the past 20 frames, which are spread over long distances due to their motion over frames.

\Figure[]()[page=7, width=0.99\linewidth]{figures/fusion-models.pdf}
{A high-level block diagram of common lidar-radar early fusion models. These models learn BEV features from both sensors, typically using similar backbones, and follow by fusing these features for joint inference. \label{fig:3d-fusion}}

 
\subsubsection{Lidar-Radar Early Fusion} 
Radars provide excellent complementary features to lidars for early fusion. Radars can enhance the accurate 3D detections provided by lidars due to their strong range and velocity features and robustness to adverse weather. Moreover, radars can improve the detection of large agents, such as buses and trucks, as radars receive strong signals from these objects. Furthermore, an early fusion model can leverage the complementary strengths of lidars and radars to reduce ghost detections that arise from each other's spurious observations. These spurious observations are often sensor-specific characteristic, such as radar clutter arising from multipath phenomena and lidar distractors arising from conditions like steam, fog, and retro-reflections. 

Interestingly, the lidar-radar pair is more compatible with early fusion compared to camera-radar or camera-lidar pairs due to the high similarity in their data. Both sensors generate point cloud data that can be represented in a BEV, and being active sensors, their data remains unaffected by lighting conditions. Nevertheless, research in the area of lidar-radar fusion remains limited. This scarcity likely arises from the limited use of lidars outside of the autonomous driving field because of their high cost, and a prevalent interest in camera-lidar fusion within the autonomous driving field due to the availability of high-quality public datasets and fewer data-related challenges, a common theme with radar data.

LiRaNet~\cite{Shah2020LiRaNetET} is a lidar-radar early fusion study that extracts spatio-temporal BEV features from raw radar data and fuses them with lidar and road map network BEV features. By incorporating radar features in the model, it not only enhances detection performance but also improves the prediction of tracks with high accelerations and at long ranges.

MVDNet~\cite{Qian2021RobustMV} is a transformer-based early fusion model that generates common region proposal tokens from lidar and radar BEV features and fuses them via a transformer cross-attention mechanism. It shows robustness on foggy weather data and overall improved detection performance over lidar-only models.  However, it is trained on a small dataset and its prediction tasks are relatively simple. ST-MVDNet~\cite{Li2022ModalityAgnosticLF} further improves MVDNet's performance on foggy weather data by implementing an envelope student-teacher model over the MVDNet model and applying a modality dropout training. 

Bi-LRFusion\cite{Wang2023BiLRFusionBL} is a recent transformer-based lidar-radar early fusion model that employs an architecture for bi-directional fusion of lidar and radar features that achieves state-of-the-art performance in the detection of dynamic objects. Bi-LRFusion enriches radar features, learned from sparse and low-resolution radar data, through cross-attention with lidar features, which are learned from dense and high-resolution lidar data. Subsequently, the model implements channel-wise concatenation of both features in a lidar-centric detection network to infer 3D bounding boxes for dynamic objects. 

While these studies have achieved enhanced detection performance through lidar-radar fusion, they have yet to fully leverage the capabilities of radars. This limitation primarily arises from the sparsity and low resolution of 3D radar data, coupled with a lack of scene diversity. Future research could emphasize the use of 4D radar-lidar datasets that encompass a wide variety of scenes, considering factors such as weather conditions, object types, and scene dynamics. 



\Figure[]()[page=1, width=\textwidth]{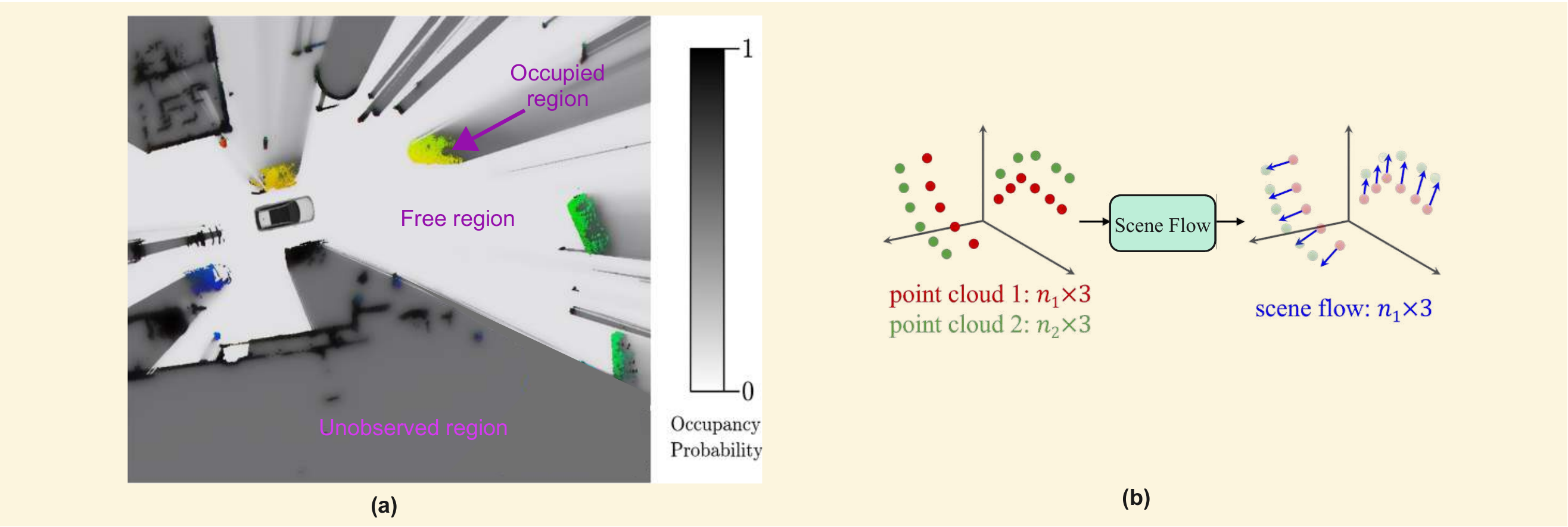}
{These two sub-figures illustrate the tasks of occupancy estimation and scene flow models. (a) Occupancy estimation models segment the scene into occupied, free, occluded, and unobserved regions~\cite{Rexin2019FusionOO}. Some models also segment regions into static and dynamic and predict the velocity of each cell. (b) Scene flow models predict a motion field to estimate the point cloud in the next frame from the point cloud in the current frame. \label{fig:occupancy}}

    
\section{Radars for occupancy-based tracking}
\label{sec:occupancy}

Occupancy estimation methods partition the ego-centric world into a fine-grained 2D or 3D grid and predict the occupancy and dynamics of each cell. These methods essentially perform an inverse sensor modeling (ISM) of the scene based on sensor data. Such modeling is typically achieved via semantic segmentation models that commonly segment the grid into occupied, free, unobserved, and ignored regions. Some models also segment the grid into static and dynamic regions and estimate the velocity of each cell. A related method, \textit{scene flow estimation}, predicts the motion field to estimate the sensor data in the next frame from the current one. Radar-based models in both methods aim to harness the unique features of radar data, including accurate velocity and range values, early detection abilities, and weather tolerance. We review the literature on both methods below.
 
\subsection{Occupancy estimation}
The early occupancy estimation methods, including those based on radar, utilized classical approaches such as Bayesian filtering, particle filtering, and Monte-Carlo to predict the occupancy of each cell~\cite{thrun2006probabilistic}. However, these traditional methods used rigid models that struggled to generalize well in the complex scenes encountered during autonomous navigation tasks.

Certain learned radar occupancy estimation models, such as~\cite{Sless2019RoadSU, Popov2022NVRadarNetRR}, utilize UNet's~\cite{Ronneberger2015UNetCN} encoder-decoder architecture on radar point clouds for 2D BEV semantic segmentation of the scene. However, these models only predict static occupancy despite having radar data containing velocity features. Moreover, their segmentation is coarse due to high angular uncertainty associated with the data. \cite{Weston2018ProbablyUD} improves the segmentation performance over these models by leveraging the uncertainty in radar data to learn the variance of occupancy for each cell along with its mean value. 

By contrast, \cite{Diehl2020RadarbasedDO} implements an evidential grid-based tracking on the radar point cloud and employs a clustering algorithm to predict dynamic occupancy. This study further shows that radar-based occupancy estimation achieves superior segmentation of static and dynamic regions compared to similar lidar-based methods, owing to a more consistent velocity grid prediction from radar data. 

The aforementioned models estimate occupancy solely using radar data and typically generate self-supervised segmentation labels from the accompanying lidar point cloud, thereby removing the need for labeled data. While these models exhibit good capabilities in segmenting dynamic regions of a scene, they suffer from coarse resolution in their occupancy estimation. To counteract this, multi-sensor occupancy models fuse lidar and radar point clouds, resulting in accurate scene occupancy and velocity estimations. 
 
As a multi-sensor occupancy model, \cite{Hoermann2017DynamicOG} predicts future dynamic occupancy of cells in the BEV from statistically computed multi-sensor dynamic occupancy input using a convolutional model. While this model outperforms single-sensor occupancy models, it occasionally mis-predicts static regions in the subsequent frame as dynamic and suffers from temporal inconsistency of tracks, particularly in occlusions.

Another multi-sensor occupancy model is \cite{Engel2018DeepOT}, which addresses the temporal inconsistency problem of \cite{Hoermann2017DynamicOG} by embedding LSTM cells into the convolutional layers. The resulting Recurrent Neural Network (RNN) architecture accumulates long-term sequence information, which adds context to dynamic tracks, enabling tracking them in occlusions and estimating their shapes more accurately. 
 
Apart from radar-only and multi-sensor models, a notable model worth mentioning is \cite{Schreiber2022AMR}, which employs a convoluational-recurrent UNet architecture to predict occupancy, velocity estimates, object segmentation, and drivable area from lidar point cloud data. It uses object labels from the dataset and the accompanying lidar point cloud to label the occupancy of the grid and perform frame interpolation for labeling the velocity of dynamic cells. This model achieves robust performance for occupancy and drivable area estimation and could potentially also be used for radar-only and multi-sensor occupancy estimation.
 
 \Figure[]()[page=2, width=\textwidth]{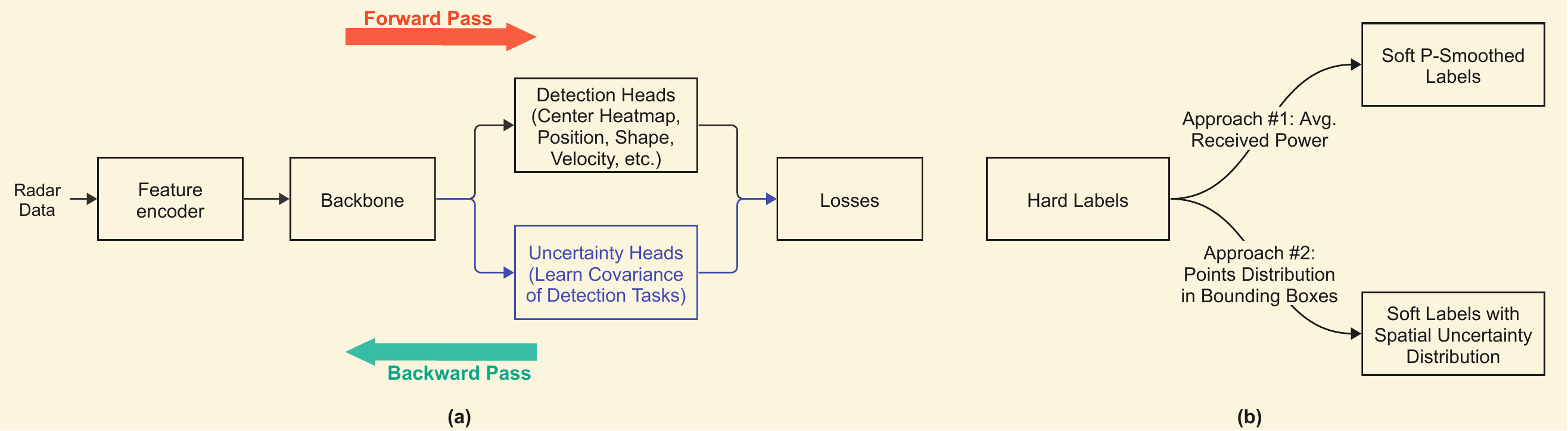}
{The high uncertainty of radar data and measurements results in less reliable detections from learned radar models. (a) Some studies seek to improve task robustness by learning the uncertainty associated with each task using auxiliary heads during the forward pass, followed by weighing the loss for each task within the total loss based on its learned uncertainty. (b) Other studies focus on improving model robustness by reducing the uncertainty of the labels. \label{fig:uncertainty}}

Several challenges remain in radar-based occupancy estimation, primarily due to the sparsity and low resolution of radar data, which result in coarse occupancy prediction. However, as some studies above have demonstrated, radar data still holds the potential for improving velocity estimation. Furthermore, radar data can enhance the weather tolerance of occupancy estimation and provide early tracking of moving agents in occlusion and over long ranges. Achieving these improvements would require (i) 4D radar data from new datasets which significantly enhance resolution and density of data over 3D radar data, and (ii) more effective multi-sensor occupancy estimation architectures that strategically leverage the strengths of radar data. 

\subsection{Scene flow}
Scene flow estimation methods predict a motion field that describes the translation of points induced by motion of dynamic agents in the scene and motion of the ego vehicle itself. Scene flow gained popularity for lidar data due to its dense point cloud representation. Some prominent lidar-based scene flow estimation models include Object Scene Flow~\cite{Menze2017ObjectSF}, Just Go with the Flow~\cite{Mittal2019JustGW}, Flot~\cite{Puy2020FLOTSF}, SLIM~\cite{Baur2021SLIMSL}, PointFlowNet~\cite{Behl2018PointFlowNetLR}, and FlowNet3D~\cite{Liu2018FlowNet3DLS}.
 
An example of a radar-based scene flow model is RaFlow~\cite{Ding2022SelfSupervisedSF}, which proposes a self-supervised model to estimate scene flow on 4D radar point cloud data. Its tasks include estimation of the motion field, static and dynamic segmentation, and rigid ego motion transformation to predict the subsequent point cloud frame. RaFlow takes two consecutive point cloud frames as input to generate a static mask and then uses this mask to segment the dynamic points. One of RaFlow's main limitations is its emphasis on scene flow for static points to improve overall metrics, which leads to less reliable flow estimation of dynamic points that carry more useful information. 

CMFlow~\cite{Ding2023HiddenG4} extends RaFlow by implementing cross-model supervised scene flow estimation using co-located data from lidar, camera, and odometer sensors. Such sensor data is incorporated at various stages of RaFlow to enhance supervision and constrain losses, thereby improving scene flow estimation of dynamic points.  
 

\section{Modeling uncertainty in radar}
\label{sec:uncertainty}

Every sensor measurement inherently carries a degree of uncertainty. The nature of this uncertainty can be either \textit{homoscedastic} or \textit{heteroscedastic}, contingent upon whether successive measurements are drawn from a random distribution with a fixed or variable variance, respectively. Inaccurately presuming heteroscedastic variables as homoscedastic can result in biased error estimates~\cite{wikipediaHomoscedasticityHeteroscedasticity}.
 
The autonomous environment is a quintessential example of heteroscedasticity due to its ever-changing dynamics. For example, measurements of a car at 10~m range typically exhibit lower variance compared to measurements of a car at 100~m. Similarly, the variance in measurements differs between a car moving straight directly in front of the ego vehicle and a car turning towards the ego vehicle due to two distinct sources of randomness.
 
Uncertainty can also be categorized as \textit{epistemic} (knowledge uncertainty) or \textit{aleatoric} (uncertainty related to inherent randomness). Epistemic uncertainty can be reduced by improving system knowledge, whereas aleatoric uncertainty, which arises from randomness inherent to the system, is irreducible and can only be lessened by selecting a new system with lower aleatoric uncertainty~\cite{Gal2016UncertaintyID}.
 
In the context of deep learning, epistemic uncertainty arises from uncertainty in the model parameters and can be mitigated by increasing the model's knowledge. This could involve selecting a larger, more balanced dataset and/or using an ensemble of models. Aleatoric uncertainty, on the other hand, pertains to the observational noise in the input data. Reducing this form of uncertainty requires improving the system, for example by using new measurement sensors with higher resolution and/or better dataset labels.
 
Radar data is often characterized by high \textit{heteroscedastic aleatoric uncertainty} due to its lower spatial resolution and weaker optical features compared to lidar and camera data. Such uncertainty often results in decreased radar model performance. However, a few studies have managed to use knowledge of uncertainty to reduce model output uncertainty or label uncertainty, thus improvin model performance. While some of these studies focus on lidar data, they can be applied to radar data as well. Important examples are reviewed below.
 
\subsection{Reducing uncertainty in model tasks}
While not a lot can be done to reduce uncertainty in sensor data post-capture, the model's performance on the data can be improved by properly leveraging data uncertainty. One radar model that demonstrates this is~\cite{Weston2018ProbablyUD}, which incorporates heteroscedastic uncertainty into its model formulation and learns the variance of occupancy for each cell in addition to its mean value. This improves segmentation of occluded regions during occupancy estimation tasks.
 
Similarly, the lidar-based work in~\cite{Feng2018LeveragingHA} leverages observation noise in the data to predict heteroscedastic aleatoric uncertainty associated with each detection, resulting in a $9\%$ improvement in average precision over similar state-of-the-art models. The model uses a modified Faster-RCNN~\cite{Ren2015FasterRT} architecture for object detection and includes auxiliary heads to predict variances in uncertainty for the region proposal network and for location and orientation prediction heads. The predicted variances regularize losses in a multi-loss function, thus promoting higher learning from informative samples and lower learning from noisy samples. 

Lastly, the vision segmentation model in~\cite{Kendall2017MultitaskLU} learns homoscedastic uncertainty for tasks and weighs each task's loss in the multi-task loss function according to its homoscedastic uncertainty, leading to improved semantic segmentation, instance segmentation, and per-pixel depth regression. 

\Figure[]()[ width=\textwidth]{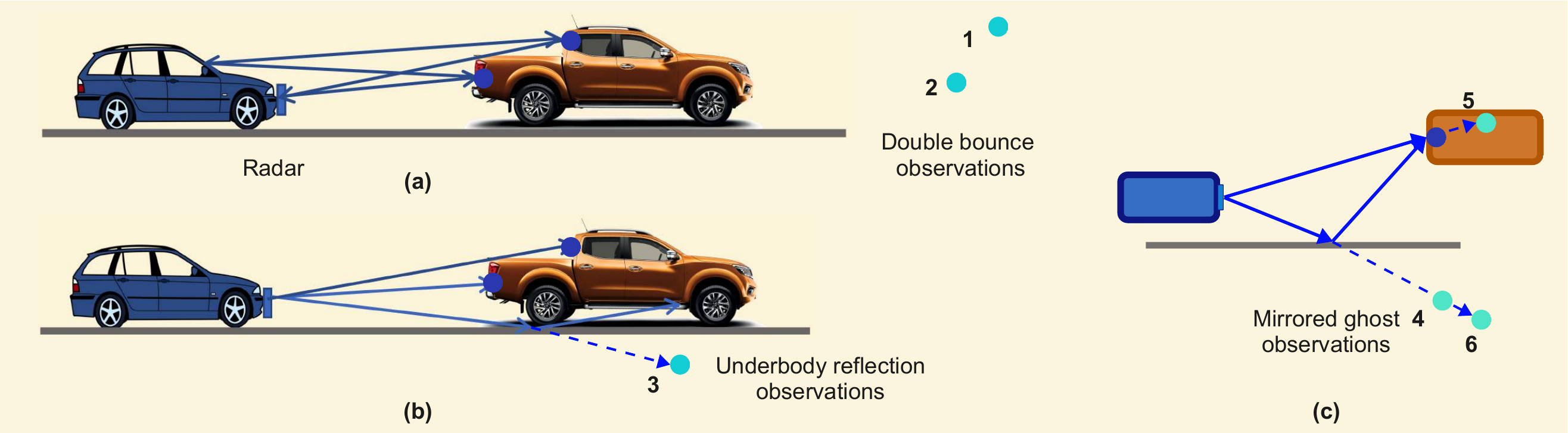}
{Three types of multipath phenomena that lead to spurious radar observations. (a) Double bounce, which arises due to two back-and-forth reflections between an object and the radar, resulting in observations roughly at double the distance and velocity, as illustrated by points $1$ and $2$. (b) Underbody reflections, which occur between the undercarriage of a vehicle and the road, depicted by point $3$. (c) Mirrored ghost detections, which occur from a surface adjacent to the main object and the radar and result in observations at various positions based on their direction of arrival and path length, shown by points $4$, $5$, and $6$. \label{fig:multipaths}}


 \subsection{Reducing uncertainty in labels}
Radar data labels often carry substantial uncertainty, which can negatively affect the performance of radar models. This uncertainty primarily originates from two sources: the lack of confidence scores and the presence of error-prone labels. All objects in the dataset carry the same confidence, irrespective of their distance, size, or visibility, which can lead to overconfident predictions for agents with low confidence. Moreover, labels for radar data are usually transferred from the spatial labels of lidar and camera data, which can be error-prone due to different imaging physics between radars and optical sensors. Such uncertainty can result in occasional erroneous predictions.

Patel et al.~\cite{Patel2021ImprovingUO} propose a solution to the confidence issue in form of a P-smoothing method that generates soft labels for objects based on their average received power. According to the radar equation~\cite{radar_eqn}, objects with lower reflectivity, such as pedestrians and bicyclists, and at longer ranges have larger aleatoric uncertainty, which is reflected in their average received power. The study trained a model on P-smoothed labels which achieved significant improvement in detection performance on a corrupted test dataset containing low-certainty agents, compared to a similar model trained on hard labels.



\section{Radar: challenges and opportunities}
\label{sec:challenges}

\subsection{Multipath and clutter}
Radars exhibit robustness against weather conditions due to their substantially larger wavelength compared to optical sensors like lidars and cameras. However, this large wavelength causes nearby surfaces, such as vehicle or guardrails, to behave as sources of specular reflections (i.e., mirrors), thus producing spurious observations in radar data known as \emph{clutter}. Thus, clutter does not arise from real objects but from indirect paths created by multiple reflections from surfaces, a phenomenon often referred to as multipath. Observations of clutter can sometimes be indistinguishable from those of real objects, leading to false positive detections in learned radar models. 

Multipath effects for autonomous vehicles can be classified into three types: double bounce, underbody reflection, and mirrored ghost detections~\cite{Kopp2021FastRC}. Radar double bounces occur due to two back-and-forth reflections between an object and the radar-equipped ego vehicle, resulting in false radar observations at double the range and velocity relative to the real object. 
Underbody reflections usually occur under a vehicle due to multiple reflections between undercarriage of the vehicle and the road. Although underbody observations increase the density of radar points on vehicles, they often appear behind the vehicle due to their longer indirect path, elongating the vehicle's shape in the learned model. Finally, mirrored ghost detections are caused by radar-reflective surfaces in the environment, such as concrete walls, guardrails, and vehicle walls. These create ghost observations that appear behind these surfaces, potentially leading to false positive detections in radar models. 

It is desirable to remove clutter from radar data to reduce false positive detections in learned models. Classical studies focused on rule-based filters to remove clutter. For example, Kopp et al.~\cite{Kopp2021FastRC} developed rule-based filters to remove three types of multipath clutter from radar data in three successive steps. However, such designs are not robust in detecting clutter in radar data from autonomous navigation scenes due to the complex nature of multipath propagation.

More recent work has focused on learned approaches for clutter filtering. In particular, architectures based on PointNet~\cite{Qi2016PointNetDL} and PointNet++~\cite{Qi2017PointNetDH} have proven to be popular~\cite{Kopp2023TacklingCI, Griebel2021AnomalyDI, Kraus2020UsingML, Chamseddine2021GhostTD}. These models learn individual point features and global scene features to predict clutter points. Among these models, \cite{Kopp2023TacklingCI} accumulates points over multiple frames to create dense point cloud data and feeds it into a custom-modified PointNet++ model. It also generates an auto-labeled dynamic clutter points dataset by filtering dynamic radar points that fall outside bounding boxes beyond a certain error margin. By contrast, \cite{Chamseddine2021GhostTD} uses a PointNet model and encodes point features in a spherical coordinate system which better learns connection between points, thereby improving the segmentation of clutter points. 

However, many clutter points, especially mirrored ghost detections, show inconsistency in motion between frames due to irregular changes in multipath length. Single-frame models like the ones mentioned above often miss these points because they lack the temporal context of the data. MATLAB's proposed multipath filter~\cite{matlab_tutorial} utilizes spatio-temporal information in radar data to achieve robust clutter detection performance. It first detects reflector surfaces in the scene from static radar points and identifies occluded dynamic clutter points arising from them. Next, it identifies occluded clutter points arising from agents detected by a learned radar model. Finally, it filters out clutter points with velocities that are inconsistent with the actual motion of the target by using a sophisticated probability hypothesis density tracker. This approach correctly detected $97\%$ of static and $80\%$ of dynamic clutter in test data. 

\Figure[]()[page=2, width=0.99\linewidth]{figures/occupancy.pdf}
{
This figure illustrates a GAN \cite{Weston2020ThereAB} that generates synthetic radar data (b) from input lidar elevation map (a). The network implements a bi-directional mapping by additionally learning to produce realistic lidar data from real radar data, which enforces cyclic consistency (depicted with dashed lines), resulting in improved quality of the generated radar data. \label{fig:sim}}

\subsection{Radar simulation}
Perception deep learning models typically require a large amount of labeled data, with a good diversity balance, for robust performance across various driving scenarios. However, both collecting such diverse data and labeling it are extremely expensive and time-consuming processes. Synthetic data, generated in simulation, offers a cost-effective way to complement real-word data.

While the synthetic data for cameras and lidars can be made realistic with modern techniques like ray-tracing and deep learning, generating realistic synthetic radar data is a challenging task. This is because radar physics is significantly different from the optical physics for which most simulation models are developed. Physics-based simulation of the radar data of a scene requires modeling of the fine-grained radar reflectivity of every object in the scene, which is difficult for complex and dynamic autonomous driving scenes.  Therefore, there are not many useful simulation models for radars.

Generative adversarial networks (GANs)~\cite{goodfellow2014generative} offer a promising solution to generate high-quality simulated radar data, especially when large amounts of real-world radar data is available. GANs operate within a framework in which a generator model produces synthetic output, while an adversarial model, trained on real world data, detects whether the output is realistic or not. When the generator model becomes so good that its output can't be reliably distinguished by the adversarial model, the adversarial model is discarded, and the generator model is used to produce synthetic data. 

One of the practical examples of GAN's potential is presented in \cite{Weston2020ThereAB}, which illustrates a GAN model that simulates real world range-azimuth heatmap radar data. The generator produces synthetic radar data from lidar elevation maps, which is a unique approach that takes advantage of the abundant and relatively easy-to-generate lidar data. Additionally, a backward mapping process is also implemented, in which real radar data is used to generate synthetic lidar elevation maps. This not only enforces cyclic consistency in the model but also ensures that the generator learns more effectively from the training data, thus leading to more realistic synthetic radar data generation. The effectiveness of GANs in generating radar data is illustrated in Fig.~\ref{fig:sim}, which shows the comparable quality of the generated synthetic data (b) and the real radar data (c). This study also demonstrated that a radar model trained on the synthetic data performs comparably to a model trained on real-world data on a real-world test data. This finding underscores the significant potential of GAN in generating synthetic radar data, as it provides a reliable method for augmenting the existing real-world datasets, thus reducing the need for expensive and time-consuming data collection and labeling.

\subsection{Velocity labeling}
Radars used in autonomous navigation, whether 3D and 4D, offer very high-resolution imaging along their velocity dimension. However, the velocity dimension of radar data is generally not labeled, as the labels for radar data are usually transferred from spatial labels of lidar and camera data. Given that radars offer relatively low resolution imaging along their azimuth and elevation dimensions, radar observations from neighboring objects can often overlap, potentially corrupting shape, velocity, and class predictions of objects in radar models. This issue could negatively impact the detection of critical objects in occlusion and at long ranges, where such objects have lower angular separation from adjacent objects.

Notably, observations from neighboring objects are clearly separated along the velocity dimension due to differences in their velocities and the fine-grained velocity imaging provided by radars. Therefore, observations from a moving agent in occlusion or at long range are easily distinguished from those of stationary objects in the scene along the velocity dimension. Thus, labeling this dimension could significantly improve the robustness of radar detection models. 

Velocity dimension labeling of radar data could be achieved by using semi-supervised or clustering-based methods. More learned approaches could also be utilized, such as the method described in~\cite{Grimm2020WarpingOR}, which generates range-Doppler labels of objects in radar data by warping a range-Doppler spectrogram into the image domain, obtaining object segmentation using camera and lidar data, and then warping it back into the range-Doppler domain. 

\section{Conclusion}
\label{sec:discussion}
In conclusion, the role of radar as a critical sensor for secure and reliable autonomous navigation has been highlighted. The notable strengths of radar include its ability to perform high-resolution velocity imaging, detect objects in occlusion and over long ranges, and maintain robust performance in adverse weather conditions. However, these strengths are counterbalanced by challenges, notably the low resolution, sparsity, clutter, and high uncertainty of radar data. This review identified and discussed key areas of focus within radar deep learning, presenting a comprehensive examination of the existing research topics, the challenges, and the potential opportunities. Along with presenting radar fundamentals, this review also included an in-depth exploration of approaches such as early and late fusion, occupancy flow estimation, uncertainty modeling, and multipath detection. 

Autonomous driving is arguably one of the most challenging applications of high-resolution radars. However, the ideas and methods discussed in this review are not limited to this application. In fact, they can be readily extended to a variety of related applications, including indoor autonomy~\cite{sun2017faster, shuai2021millieye}, ground penetrating radar~\cite{Srivastav2020AHD, 8441665, giovanneschi2022modern}, next-generation planetary rovers, and joint communication and sensing for future wireless networks. 

\nocite{*}
\bibliographystyle{ieeetr}
\bibliography{access}

\begin{IEEEbiography}[{\includegraphics[width=1in,height=1.25in,clip,keepaspectratio]{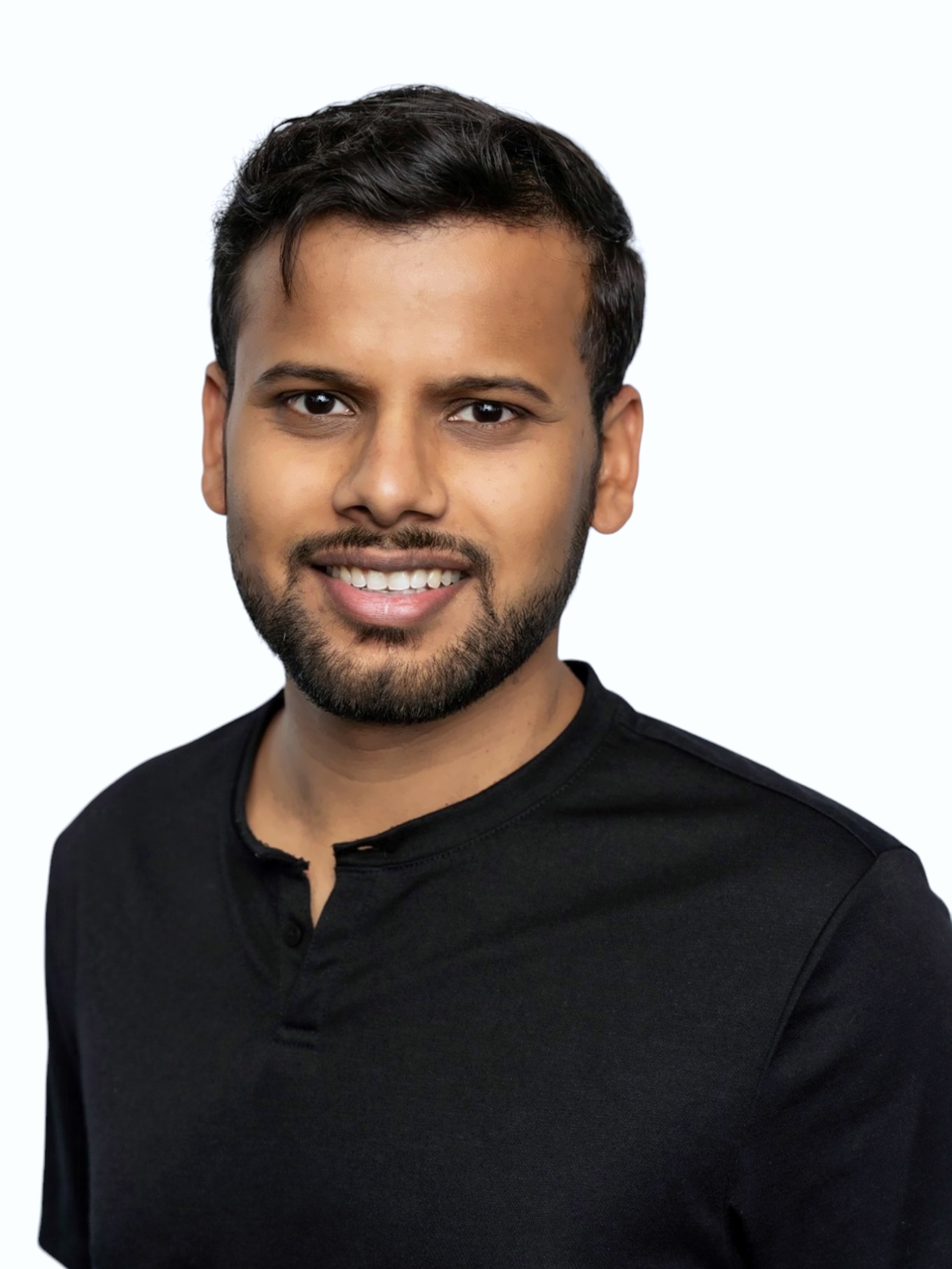}}]{Arvind Srivastav} is currently a Software Engineer at Zoox, Inc., where he leads research and development projects related to radar perception. He received his B.Tech. degree in Electronics and Electrical Communications Engineering from the Indian Institute of Technology, Kharagpur, West Bengal, India, in 2019, and the M.S degree in Electrical Engineering from Stanford University, Stanford, CA, USA, in 2021. He was a Graduate Research Assistant in the Arbabian Lab at Stanford from 2019 to 2021. His interests include radar deep learning, early sensor fusion, and autonomous perception. He has 6 publications in peer-reviewed journals and conferences and has been awarded 7 patents. He is a recipient of Innovation Scholar Award from the President of India in 2016.   
\end{IEEEbiography}

\begin{IEEEbiography}[{\includegraphics[width=1in,height=1.25in,clip,keepaspectratio]{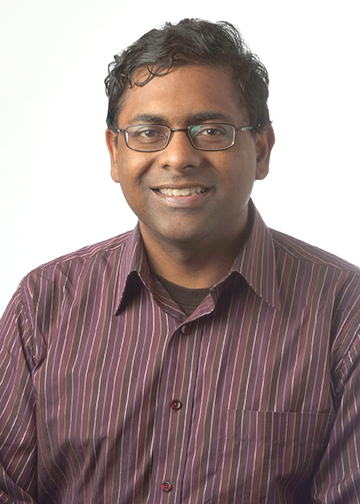}}]{Soumyajit Mandal} (Senior Member, IEEE) received the B.Tech. degree in electronics and electrical communications engineering from the Indian Institute of Technology, Kharagpur, West Bengal, India, in 2002, and the S.M. and Ph.D. degrees in electrical engineering from the Massachusetts Institute of Technology, Cambridge, MA, USA, in 2004 and 2009, respectively. He was a Research Scientist at Schlumberger-Doll Research, Cambridge, from 2010 to 2014, an Assistant Professor at the Department of Electrical Engineering and Computer Science, Case Western Reserve University, Cleveland, OH, USA, from 2014 to 2019, and an Associate Professor at the Department of Electrical and Computer Engineering, University of Florida, Gainesville, FL, USA, from 2019 to 2021. He is currently a Research Staff Member with the Instrumentation Division at Brookhaven National Laboratory, Upton, NY, USA. He has over 175 publications in peer-reviewed journals and conferences and has been awarded 26 patents. His research interests include analog and biological computation, magnetic resonance sensors, low-power analog and RF circuits, and precision instrumentation for various biomedical and sensor interface applications. He was a Recipient of the President of India Gold Medal, in 2002; the MIT Microsystems Technology Laboratories (MTL) Doctoral Dissertation Award, in 2009; the T. Keith Glennan Fellowship, in 2016; and the IIT Kharagpur Young Alumni Achiever Award, in 2018.
\end{IEEEbiography}

\EOD

\end{document}